\documentclass[sigconf]{acmart}

\usepackage[ruled,vlined]{algorithm2e}
\usepackage{graphicx}
\usepackage{multirow}
\usepackage{flushend}
\usepackage{balance}
\AtBeginDocument{%
  }

\setcopyright{acmcopyright}
\copyrightyear{2018}
\acmYear{2018}
\acmDOI{XXXXXXX.XXXXXXX}

\begin{document}

\title{FinSQL: Model-Agnostic LLMs-based Text-to-SQL Framework for Financial Analysis}

\settopmatter{authorsperrow=4}

\author{Chao Zhang}
\affiliation{%
  \institution{Zhejiang University}
  \city{}
  \country{}
}
\email{zjuzhangchao@zju.edu.cn}

\author{Yuren Mao}
\affiliation{%
  \institution{Zhejiang University}
  \city{}
  \country{}
}
\email{yuren.mao@zju.edu.cn}

\author{Yijiang Fan}
\affiliation{%
  \institution{Zhejiang University}
  \city{}
  \country{}
}
\email{yijiangfan@zju.edu.cn}

\author{Yu Mi}
\affiliation{%
  \institution{Zhejiang University}
  \city{}
  \country{}
}
\email{miyu@zju.edu.cn}

\author{Yunjun Gao}
\affiliation{%
  \institution{Zhejiang University}
  \city{}
  \country{}
}
\email{gaoyj@zju.edu.cn}

\author{Lu Chen}
\affiliation{%
  \institution{Zhejiang University}
  \city{}
  \country{}
}
\email{luchen@zju.edu.cn}

\author{Dongfang Lou}
\affiliation{%
  \institution{Hundsun Technologies INC.}
  \city{}
  \country{}
}
\email{loudongfang2022@gmail.com}

\author{Jinshu Lin}
\affiliation{%
  \institution{Hundsun Technologies INC.}
  \city{}
  \country{}
}
\email{linjs13607@hundsun.com}

\renewcommand{\shortauthors}{Chao Zhang et al.}

\begin{abstract}
Text-to-SQL, which provides zero-code interface for operating relational databases, has gained much attention in  financial analysis; because, financial professionals may not well-skilled in SQL programming. However, until now, there is no practical Text-to-SQL benchmark dataset for financial analysis, and existing Text-to-SQL methods have not considered the unique characteristics of databases in financial applications, such as commonly existing wide tables. To address these issues, we collect a practical Text-to-SQL benchmark dataset  and propose a model-agnostic Large Language Model (LLMs)-based Text-to-SQL framework for financial analysis. The benchmark dataset, BULL, is collected from the practical financial analysis business of Hundsun Technologies Inc., including databases for fund, stock, and macro economy. Besides, the proposed LLMs-based Text-to-SQL framework, FinSQL, provides a systematic treatment for financial Text-to-SQL from the perspectives of prompt construction, parameter-efficient fine-tuning and output calibration. 
Extensive experimental results on BULL demonstrate that FinSQL achieves the state-of-the-art Text-to-SQL performance at a small cost; 
furthermore, FinSQL can bring up to 36.64\% performance improvement in scenarios requiring few-shot cross-database model transfer. 
\end{abstract}



\maketitle

\section{Introduction}
Text-to-SQL aims to transform natural language questions into executable SQL queries, which enables low-code operations for relational databases. It can facilitate the data access procedure for non-professional database users who are not familiar with SQL and has gained much attention in various areas, especially in financial analysis. While financial professionals (e.g., investment advisors) need to query relevant databases frequently, they are usually not well-skilled in SQL programming. Therefore, Text-to-SQL is significantly important for financial analysis and has  gained much attention. 

\vspace {0.1cm}

However, there is no Text-to-SQL benchmark dataset for financial analysis, and existing Text-to-SQL methods have not considered the unique characteristics of databases used in  financial analysis. To address these issues, we construct a practical Text-to-SQL dataset for financial analysis based on the intelligent investment assistant product of Hundsun Technologies Inc., which facilitates more than 50 financial institutions (including Alipay, China Merchants Bank, and so on) and serves millions of personal users. This dataset, dubbed BULL, contains three databases corresponding to fund, stock, and macro economy respectively. Besides, in this dataset, there are 4,966  natural language question-SQL pairs annotated by financial professionals, data scientists, and software engineers from Hundsun Technologies Inc. Furthermore, BULL has both English and Chinese versions.


  \vspace {0.1cm}
        
\begin{figure*}[tbp]
  \vspace {0.1cm}
  \centering
  \includegraphics[width=\textwidth]{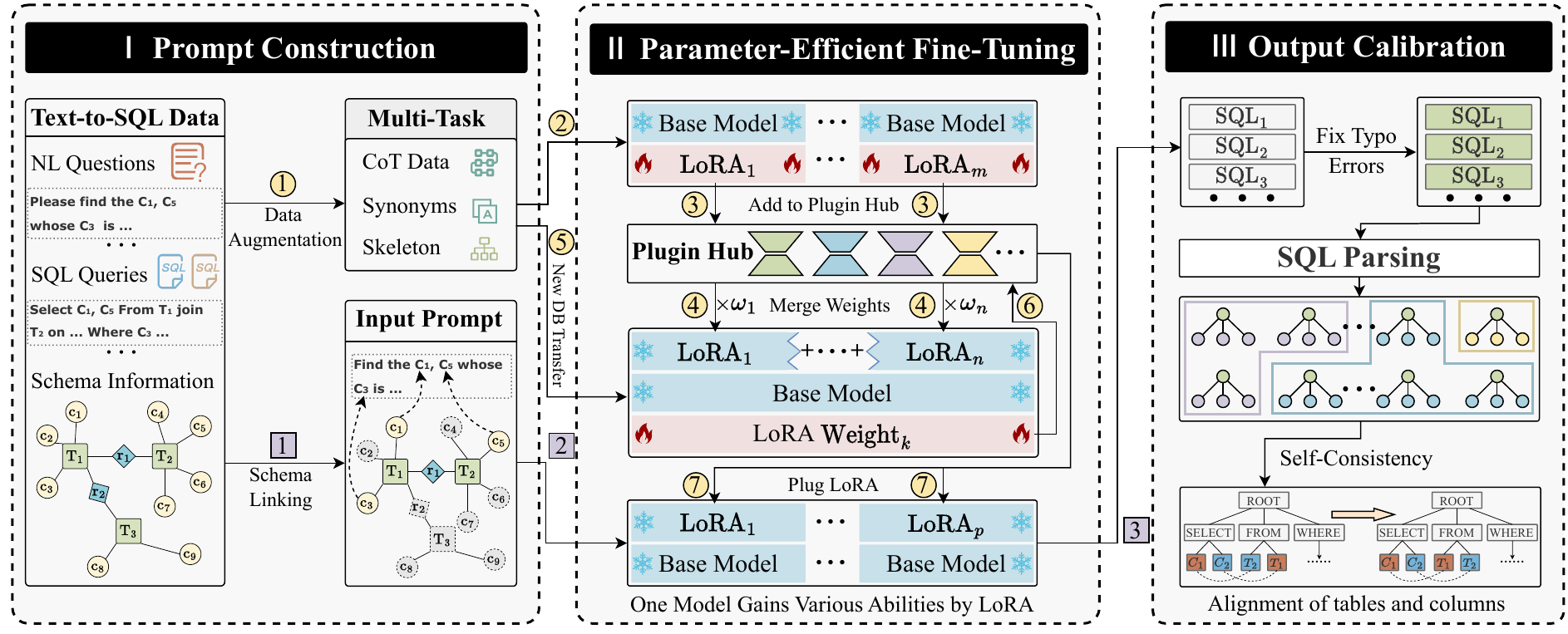}
  \caption{The overview of FinSQL framework.
  In the training stage, the training data is first augmented with a hybrid data augmentation method. Then, the augmented data is used to train LoRA plugins which are used to handle various Text-to-SQL tasks. The LoRA plugins are managed by a LoRA plugin hub. In the inference stage, schema linking is firstly conducted to obtain concise prompt, and then the prompt inputs into a LLM model which consists of a base model and a LoRA module constructed with merged LoRA plugins. Finally, the output of the LLM model is calibrated to ensure the correctness of the model output.
   The ice~\includegraphics[width=0.35cm]{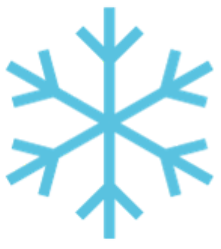} and fire~\includegraphics[width=0.35cm]{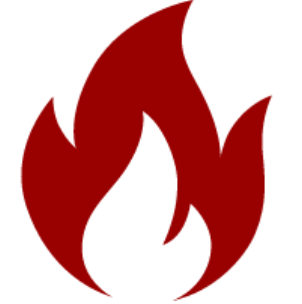} in the picture mean freezing and updating model weights respectively. The number in~\includegraphics[width=0.3cm]{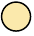} and ~\includegraphics[width=0.3cm]{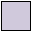} indicate the process step of training and inference respectively.}
  \label{fig:overview}
\end{figure*}

Compared with the widely used Text-to-SQL benchmark datasets (e.g., Spider\cite{spider} and BIRD \cite{bird}), BULL has much more tables for each database and much more columns for each table, illustrating as in Table \ref{tab:dataset comparision}. Furthermore, table and column names in BULL are often expressed with abbreviations or vague representations. These characters require financial Text-to-SQL models to support large input context length and have strong context understanding ability. Fortunately, Large Language Models (LLMs)-based Text-to-SQL can satisfy these requirements, and several LLMs-based Text-to-SQL methods have been proposed recently. However, existing state-of-the-art LLMs-based Text-to-SQL methods typically depend on OpenAI's APIs, such as GPT-3.5-turbo or GPT-4, which are expensive and have risks of information leakage. Therefore, these methods cannot be used in the financial applications where the information privacy is critically important.

\vspace {0.1cm}

To avoid information leakage, a feasible way is to adopt open-source LLMs (e.g., LLaMA\cite{llama2} and Baichuan~\cite{baichuan2}) and train  them in private domains. However, it faces three challenges: 
(1) Schema linking dilemma and data scarcity.  It is difficult to establish connections between question and schema items for financial databases usually have a large  number of columns and tables. Furthermore, due to the labeling cost, the number and diversity of labeled question-SQL pairs are limited.
These issues obstruct the construction of concise and diversified prompts, which hinders the model's performance; 
(2) Resource-consuming fine-tuning and cross-database generalization difficulties.
Fine-tuning LLMs on downstream tasks demands several days of computation across multiple GPUs. The substantial cost associated with model updates and iterations poses a considerable challenge. Additionally, transferring the model to a new database incurs significant costs, which impedes the cross-database transfer.
(3) Inconsistent output. 
Due to the inherent randomness and the decoder strategy of sampling, LLMs often generate inconsistent outputs, leading to syntactically incorrect and invalid SQL queries. 

To tackle these challenges, this paper proposes a model-agnostic LLMs-based Finacial Text-to-SQL model training and inference framework, dubbed FinSQL. It can be used to develop Text-to-SQL models based on any open-source LLMs. Figure~\ref{fig:overview} demonstrates the overall overview of FinSQL, which consists of three key components: prompt construction, parameter-efficient fine-tuning, and output calibration, corresponding to the above challenges correspondingly. 
Specifically, prompt construction consists of a parallel schema linking method and  a hybrid data augmentation method, which help to construct more  concise and diverse 
prompts and enhance the model's performance from the input side. 
The parameter-efficient fine-tuning
component adopts Low-Rank Adaptation (LoRA) to  fine-tune a very small percentage of parameters (<1\%) to obtain weight plugins for different business scenarios and manages these plugins through a plugin hub. 
Based on this plugin hub, the database-specific Text-to-SQL models can achieve efficiently few-shot cross-database transfer. In
output calibration, SQL post-processing is performed to enhance the correctness of the generated SQL. These three components contribute to FinSQL's superior performance.

\begin{table}[tbp]
\setlength{\tabcolsep}{3.13mm}
  \caption{Differences Between Datasets}
  \label{tab:dataset comparision}
  \begin{tabular}{ccccc}
    \toprule
    Dataset&Example&Table/DB&Column/DB\\
    \midrule
    WikiSQL~\cite{wikisql} & 80654 & 1  & 6.3\\ 
    Spider~\cite{spider} & 10181 & 5.1  & 27.1\\ 
    BIRD~\cite{bird} & 12751 & 7.3  & 54.2\\
    \midrule
    \textbf{BULL} & \textbf{4966} & \textbf{26}  & \textbf{390} \\ 
  \bottomrule
\end{tabular}
\vspace {-0.5cm}
\end{table}

Our contributions can be summarized as follows:
\begin{itemize}
	\item We propose BULL, a practical benchmark dataset for financial Text-to-SQL. 
        \item We propose FinSQL, a model-agnostic LLMs-based Text-to-SQL framework for financial analysis. 
        \item Extensive experimental results on BULL demonstrate that FinSQL is model-agnostic and able to achieve the state-of-the-art performance; furthermore, FinSQL can bring up to 36.64\% performance improvement  in scenarios requiring few-shot cross-database model transfer.
\end{itemize}

\section{Related Works}
In this section, we introduce various Text-to-SQL datasets. We explore the distinctive features of various datasets and highlight the existing gap between these datasets and real-world financial scenarios. Additionally, we present an overview of recent developments in Text-to-SQL models.

\subsection{Text-to-SQL Datasets}
High-quality datasets play an important role in the development and evaluation of Text-to-SQL systems. Early Text-to-SQL datasets such as GeoQuery~\cite{genquery} and Scholar~\cite{scholar} provided valuable insights, but they were limited in terms of the number of queries, focused on single-database scenarios, and featured relatively simple SQL queries. WikiSQL~\cite{wikisql} and Spider~\cite{spider}, on the other hand, offer over 10,000 SQL queries, enabling cross-database migration scenarios and introducing complex multi-table queries. 

Despite these improvements achieved, there is still a considerable gap between these datasets and real-world scenarios. Databases used in real-world applications are often larger and may provide external knowledge to help users understand schema information. KaggleDBQA~\cite{kaggledbqa} created 272 SQL queries across eight databases and introduced external knowledge to explain the meaning of columns. BIRD~\cite{bird} offers a greater number of SQL query samples and larger databases and incorporates external knowledge between questions and database contents, making Text-to-SQL datasets more suitable to real-world situations. 

However, these datasets still do not fully satisfy the demands of the industry, which are often more complex and challenging. As shown in Table~\ref{tab:dataset comparision}, in real industrial scenarios, the number of tables and columns in databases far exceeds existing open-source datasets. This poses challenges for existing Text-to-SQL methods in terms of incorporating schema information, making it difficult for them to work effectively in this scenario.

\subsection{Text-to-SQL Models}
Text-to-SQL models have gained significant advancements in recent years~\cite{Text-to-SQL_survey}. Early rule-based approaches rely on handcrafted templates to generate SQL, showing some effectiveness but being heavily dependent on manually defined rules. They have limited applicability to other scenarios, lack scalability, and struggle with generalization.

To overcome these limitations, researchers develop Text-to-SQL methods based on the Seq2Seq architecture. IRNet~\cite{irnet} uses an encoder to represent questions and schema, while RAT-SQL~\cite{ratsql}, LGESQL~\cite{lgesql}, and S$^{2}$SQL~\cite{s2sql} employ graph neural networks to capture alignment relationships between questions and schema, and then use a decoder to generate SQL queries.

Subsequently, with the development of large language models, pre-trained models like T5, mT5, LLaMA, and others demonstrated enhanced language understanding and generalization capabilities. Methods based on fine-tuning these models achieved better results in Text-to-SQL tasks. Graphix~\cite{graphix-t5} enables T5 with multi-hop reasoning ability. Picard~\cite{picard} uses a constrained decoder to enhance the quality of generated SQL queries. RESDSQL~\cite{RESDSQL} uses a two-stage approach to first retrieve schema elements relevant to the question and then generate the corresponding SQL query, which is now the SOTA fine-tuning based method in Spider leaderboard. 

More recently, large language models (LLMs) have exhibited remarkable capabilities and left a deep impression. With the In-context learning technique, these models can understand human instructions and execute tasks without requiring retraining. Currently, leading approaches on the Spider leaderboard are based on GPT models using In-context Learning, such as C3, DIN-SQL, and DAIL-SQL. However, these methods involve invoking the OpenAI API, which can be costly and pose privacy risks. Furthermore, instruction templates need redesigning for different scenarios.

\section{Preliminaries}

In this section, we formalize the problem of Text-to-SQL for finance and introduce the related techniques employed in our proposed method, including data augment based on LLMs, schema linking, large language models, low rank adaption, self-consistency. 

\subsection{Problem Formulation}

This paper focuses on LLM-based Text-to-SQL systems. In this section, we first give the definition of Text-to-SQL and then introduce the LLM-based Text-to-SQL systems.

\textbf{Text-to-SQL.} Given a natural language query  $\mathcal{Q}$ and its corresponding database schema $\mathcal{S}=(\mathcal{T}, \mathcal{C}, \mathcal{R})$ where $\mathcal{T}=(t_1, t_2, ...)$, $\mathcal{C}=(c_1, c_2, ...)$, $\mathcal{R}=(r_1, r_2, ...)$, which present multiple tables, columns and foreign keys relations  respectively, a Text-to-SQL system aims to generate a executable SQL query $\mathcal{Y}$ corresponding to $\mathcal{Q}$. Traditional Text-to-SQL methods typically adopt the encoder-only or encoder-decoder models (e.g. SyntaxSQLNet~\cite{syntaxsqlnet}, LGESQL~\cite{lgesql}). Recently, benefiting from the emergent abilities of the decoder-only Large Language Models (LLMs), LLM-based Text-to-SQL methods achieve the state-of-art performance. Next, we introduce the working mechanism of the LLM-based Text-to-SQL system.

\textbf{LLM-based Text-to-SQL System.} LLM-based Text-to-SQL utilizes LLMs $\mathcal{M}$ to generate SQL queries. It begins by receiving an instruction prompt, denoted as $\mathcal{P}(\mathcal{Q},\mathcal{S})$, which is generated by combining a question $\mathcal{Q}$ and a database schema $\mathcal{S}$ into a prompt template. Subsequently, the model $\mathcal{M}$ estimates the probability distribution over SQL query $\mathcal{Y}$ and generates it token by token. The generation process for the SQL query $\mathcal{Y}$ can be formulated as follows:

\begin{equation}
    P_{\mathcal{M}}(\mathcal{Y}|\mathcal{P}(\mathcal{Q},\mathcal{S}))=\prod_{i=1}^{|\mathcal{Y}|}P_{\mathcal{M}}(\mathcal{Y}_i|\mathcal{P}(\mathcal{Q},\mathcal{S}),\mathcal{Y}<i)
\end{equation}

\subsection{Schema Linking} \label{schema_linking preliminary}

Schema linking, which aims to link the meta data of a database schema to a natural language query, is a fundamental component in the Text-to-SQL system\cite{lei2020re}. Specifically, given a  natural language query $\mathcal{Q}$ and its corresponding database schema $\mathcal{S}=(\mathcal{T}, \mathcal{C}, \mathcal{R})$, schema linking extracts a subset of meta data that is relevant  $\mathcal{Q}$. The results of schema linking is denoted as $\mathcal{S}'=(\mathcal{T}', \mathcal{C}', \mathcal{R}')$, where $\mathcal{T}' \in \mathcal{T}, \mathcal{C}' \in \mathcal{C}'$ and $\mathcal{R}' \in \mathcal{R}$. Schema linking can help to reduce LLMs' misunderstanding about the natural language query by reducing the noise caused by redundant tables and columns. Consequently, it can effectively enhance the quality of generated SQL queries.

Early Text-to-SQL methods, exemplified by IRNet and IESQL, rely on rule-based or character-matching techniques to perform exact or partial matches between the tokens in the question and the elements in the database schema. Subsequently, they identified the relationships between the query and the table and column names, which are then leveraged to generate the SQL query corresponding to the question. 
However, the generalization ability of these approaches is limited. When encountering words with similar meanings, they often fail to make accurate matches. Furthermore, these methods cannot capture the comprehensive semantic relationships between natural language and table structures, as well as the various relationships among schema elements. To express this structural relationship, some researchers employ graph neural networks (GNN) for representation, such as LGESQL, RATSQL, and others, which have yielded promising results. However, These approaches necessitate the use of specific GNN modules as part of the overall model and are difficult to adapt to prevalent open-source large models such as LLaMA, Baichuan, ChatGLM, etc. 

To enhance compatibility with LLMs, some methods isolate schema linking as a separate module. RESDSQL utilizes the Roberta model as a base model to build a Cross-Encoder model to recall and rank schema elements. C3 and DIN-SQL provide well-designed in-context learning instructions to the GPT models for retrieving relevant schema items. These methods  achieve outstanding performance on popular benchmarks, such as spider leaderboard.

\subsection{Chain of Thought} \label{CoT}
Chain of Thought (CoT)~\cite{cot} is a method that enables LLMs to think like humans to solve complex reasoning tasks. In the CoT setting, LLMs first output the reasoning progress and then provide the final answer, which demonstrates substantial improvements in handling reasoning tasks. In the challenging arithmetic reasoning benchmark GSM8K~\cite{gsm8k}, the CoT technique has led to a substantial improvement in Google's large language model, Palm-540B~\cite{palm}. The accuracy on this benchmark increases significantly, rising from 17.9\% to 56.5\%~\cite{cot}. Additionally, the CoT technique instructs large language models (LLMs) to first generate the reasoning process and then produce the final result. This process enhances the interpretability of the model's output.

\subsection{Parameter-Efficient Fine-Tuning}
Parameter-Efficient Fine-Tuning (PEFT) methods (e.g., Adapter~\cite{adapter}, Prompt Tuning~\cite{prompt_tuning}, Prefix-Tuning~\cite{prefix_tuning}, LoRA~\cite{LoRA}), which only tunes a small fraction of parameters of pre-trained LLMs and can achieve comparable performance with full-parameter fine-tuning, enables an LLM to adapt to downstream tasks with few computational resources. 
They only need to fine-tune a small fraction of the model weights (<1\%) to adapt it to downstream tasks. This process can be completed in a short time on a single GPU.

\subsection{Cost of LLMs}
Different large models have varying context lengths, typically measured in terms of the number of tokens. In OpenAI, GPT-4 offers context lengths of 8k and 32k tokens. The efficiency of the tokenizer for GPT-4 and ChatGPT is approximately such that 1000 tokens correspond to around 700 English words. As a result, the amount of information a large model can handle at once is limited. 

We can access the ChatGPT and GPT-4 models by utilizing their respective paid API interfaces. The pricing details for these models are provided in the Table~\ref{tab:price}. 
\begin{table}
\setlength{\tabcolsep}{1.5mm}
  \caption{API Price of GPT Models}
  \label{tab:price}
  \begin{tabular}{ccc}
    \toprule
    Model&Input&Output\\
    \midrule
    GPT-4-8k & \$0.03 / 1K tokens & \$0.06 / 1K tokens\\
    GPT-4-32k & \$0.06 / 1K tokens & \$0.12 / 1K tokens\\
    GPT-3.5-turbo-1106 & \$0.001 / 1K tokens & \$0.002 / 1K tokens\\
  \bottomrule
\end{tabular}
\end{table}

\subsection{Hallucination of LLMs}
The phenomenon of hallucination in LLMs refers to the situation that LLMs, when generating factual content, sometimes produce information that appears to be correct but contradicts actual facts. The essence of this phenomenon lies in the LLMs' inability to maintain precise control over knowledge \cite{zhao2023survey}. Even the most powerful LLMs like GPT-4 may still encounter this issue. 

\subsection{Uncertrainty of LLMs}
The output of LLMs exhibits uncertainty and instability due to the inherent randomness within them \cite{lin2022teaching}. LLMs generate multiple results when executed multiple times with the same input prompt. This issue may result in the generation of a correct SQL query one time, but an incorrect one the next time. Therefore, it is of great significance for Text-to-SQL to enhance the consistency of the generated output content from LLMs.

\section{Overview}
Large Language Model (LLM)-based Text-to-SQL methods for financial analysis typically encounter three challenges: 
(1) Schema linking dilemma and data scarcity. 
Databases for finance applications usually consist of a large number of wide tables that contain lots of columns, which makes it hard to establish connections between question and schema items. Furthermore, the number and  diversity of labeled question-SQL pairs are limited.
These issues obstruct the construction of concise and diversified prompts, which hinders the model's performance; 
(2) Resource-consuming fine-tuning and cross-database generalization difficulties.
Fine-tuning LLMs on downstream tasks demands several days of computation across multiple GPUs. The substantial cost associated with model updates and iterations poses a considerable challenge. Additionally, transitioning the model to new database scenarios incurs significant costs.
(3) Inconsistent output. 
Due to the inherent randomness and the decoder strategy of sampling, LLMs often generate inconsistent outputs, which leads to syntactically incorrect and invalid SQL queries. 

To tackle these challenges, we propose a Text-to-\textbf{SQL} framework for \textbf{Fin}ancial data, \textbf{FinSQL}, which consists of three key components addressing the challenges correspondingly. 
Figure~\ref{fig:overview} provides an overview of the proposed FinSQL, which consists of three key components, \textbf{Prompt Construction}, \textbf{Parameter-Efficient Fine-Tuning}, and \textbf{Output Calibration}.
The Prompt Construction module proposes a parallel Cross-Encoder model to retrieve relevant schema elements and a hybrid data augmentation method to enrich the original dataset, helping to construct clear prompts and enrich training data. The Parameter-Efficient Fine-Tuning module utilizes the augmented dataset to fine-tune LLMs with LoRA method, which can significantly reduce the computational cost. This module also provides a LoRA weight merging method for further training, which significantly improves the performance of LLMs when transferring across different databases. The Output Calibration designs an algorithm that performs SQL syntax checking and self-consistency to improve the overall consistency and coherence of the final output SQL queries. The details of these modules are introduced as follows.

\textbf{Prompt Construction}. 
This module focuses on constructing clear prompts and enriching the training data, which are beneficial to both the training and inference processes.
It augments the original data based on LLMs and well-designed rules, resulting in a new dataset that covers three SQL-related instruction prompts: chain-of-thought data, synonymous question data, and skeleton data. The augmented data is later used to fine-tune LLMs, which can significantly improve the model's reasoning and comprehensive ability in Text-to-SQL compared to the original data. Besides, this module trains a Cross-Encoder model for retrieving relevant schema elements corresponding to the question. It can help eliminate irrelevant elements and generate more concise and clear prompts, which leads to significant improvement in model performance. 

\textbf{Parameter-Efficient Fine-tuning}. 
Provided with the augmented data, this module uses the Parameter-Efficient Fine-Tuning (PEFT) methods, such as LoRA, to adapt LLMs to downstream Text-to-SQL task on a single GPU in a short time. These trained LoRA weights are then added to the LoRA plugin Hub. In the case of low-resource database scenarios, the original parameters of the base model are first updated by merging them with previous LoRA weights from Plugin Hub. Then, an additional LoRA module is applied for further fine-tuning on this foundation. This database domain transfer approach demonstrates the ability to achieve strong performance with limited data. A database corresponds to a LoRA module, which stores in the LoRA Plugin Hub. When dealing with different database scenarios, we can effortlessly plug the relevant LoRA module into the base model to generate SQL queries. 

\textbf{Output Calibration}. 
Due to the inherent randomness within LLMs, the SQL queries LLMs generate are often uncertain and may not strictly adhere to the SQL syntax rules, resulting in invalid SQL queries. This module incorporates a post-processing augmentation algorithm specifically tailored for SQL, which leads to more consistent SQL queries. It initially corrects typo errors in SQL and subsequently parses the SQL queries to extract keywords and values, which are then utilized to 
mine the most consistent SQL. At last, it employs alignment between tables and columns for the SQL queries.

\section{BULL: Financial Text-to-SQL Dataset} \label{section:dataset}
BULL consists of three databases corresponding to fund, stock, and macro economy respectively. Besides, there are 4,966  natural language questions-SQL pairs annotated by financial professionals, data scientists, and software engineers from Hundsun Technologies Inc.  In this section, we introduce the details about the databases and the process of annotation.

\begin{figure}[tbp]
  \includegraphics[width=\linewidth]{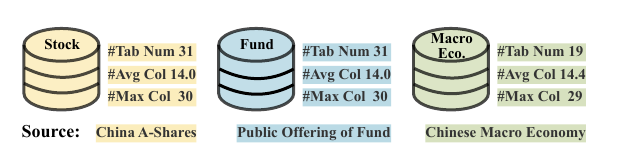}
  \vspace{-0.4cm}
  \caption{The introduction of BULL databases. The English and Chinese versions of BULL share the same database structure. \#Tab Num represents the number of tables in the database. \#Avg Col and \#Max Col mean the average and maximum number of columns in each table within the database. }
  \label{fig:database_intro}
  \vspace{-0.4cm}
\end{figure}

\subsection{Database Source}
\begin{figure}[tbp]
  \includegraphics[width=\linewidth]{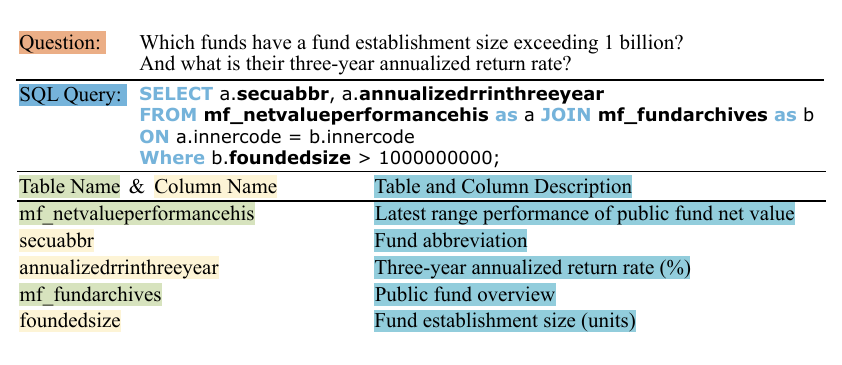}
  \vspace{-0.4cm}
  \caption{An example of BULL dataset}
  \label{fig:bull_example}
\end{figure}

BULL consists of three databases that
build based on the intelligent investment assistant product of Hundsun Technologies Inc. The three databases are constructed with data related to China A-Shares, China Public Offering of Fund, and Chinese Marco Economy. The three databases contain 31, 28, and 19 tables respectively. The cutoff date of the including data is up to April 2022.
Most of the tables have more than ten columns. Figure~\ref{fig:database_intro} illustrates the details of this dataset. Furthermore, table and column names in BULL are often expressed using abbreviations or vague representations. One example of the data in BULL is shown in Figure~\ref{fig:bull_example}.

\subsection{Annotation}

Generating question-SQL pairs for training Text-to-SQL models requires professional financial knowledge. To construct question-SQL pairs for BULL, several financial professionals, data scientists and software engineers of Hundsun Technologies Inc. work together for more than one week. The financial professionals and data scientists first write 4966 diverse Chinese questions. Then, the software engineers write corresponding SQL queries for the questions. 
Subsequently, three interns with both  knowledge of database and finance validate the correctness of these question-SQL pairs. Furthermore, they extend BULL to an English version. Specifically, they translate the descriptions of tables and columns using professional financial terminology. Then, two interns translate the questions, while the other intern scrutinizes the translated sentences for grammatical correctness and adherence to customary financial expressions. 
Finally, they rewrite the SQL queries to suit English databases. This process involves replacing the Chinese values with their English counterparts mentioned in the questions.
The above steps result in the creation of the comprehensive BULL dataset.


\section{Prompt Construction}
Prompts significantly impact the performance of LLMs-based Text-to-SQL. 
To construct proper prompts, this paper proposes a hybrid data augmentation strategy and a parallel schema linking method. 
The hybrid data augmentation strategy jointly leverages LLMs and carefully designed rules to enrich the training data. 
This process can improve both the quality and quantity of the training data, thereby improving the performance of the models trained in the subsequent steps. Besides, 
the parallel schema linking method adopts a Cross-Encoder model to retrieve several schema elements relevant to the question in parallel, which can effectively reduce the irrelevant content and noise for the input data to the model. 
It ultimately assists the model in generating SQL queries with higher quality.

\subsection{Hybrid Data Augmentation} \label{section data augmentation}
In finance applications, data labeling requires professional knowledge and numerous manual efforts, which results in a shortage of professional annotators and constrained labeled data, leading to insufficient data and a lack of data diversity.
It is necessary to enhance the diversity of the dataset because a diverse dataset reflects a broader range of real-world scenarios and conditions. 

Data augmentation is a natural choice to enrich the dataset. However, existing Text-to-SQL augmentation methods primarily focus on enhancing SQL queries and questions~\cite{Recent_advances_in_text-to-SQL}, without considering the reasoning process underlying the generation of SQL. With the development of LLMs, more and more data augmentation methods about reasoning ability based on LLMs have emerged\cite{2023arXiv230213007D}, which achieves excellent results. In this paper, we propose a hybrid data augmentation method consisting of three data forms, denoted as chain-of-thought, synonymous questions, and skeleton SQL, which correspond to the reasoning ability, meaningful expressions, and SQL structures respectively.

\subsubsection{Chain-of-thought Augmentation}
The Chain-of-Thought (CoT) prompting technique can effectively improve the reasoning ability of LLMs~\cite{cot}. Existing methods, such as CoT-KA\cite{CoT-KA}, leverage CoT prompting to instruct GPT-3 in generating CoT reasoning content. Subsequently, it synthesizes new CoT data using this content. However, these methods are unable to apply to Text-to-SQL, because LLMs are likely to generate wrong SQL queries, which can lead to incorrect CoT reason content and is harmful to the model performance. Therefore, it is vital to increase the accuracy of the generated CoT reasoning content.

\begin{figure}[tbp]
  \includegraphics[width=\linewidth]{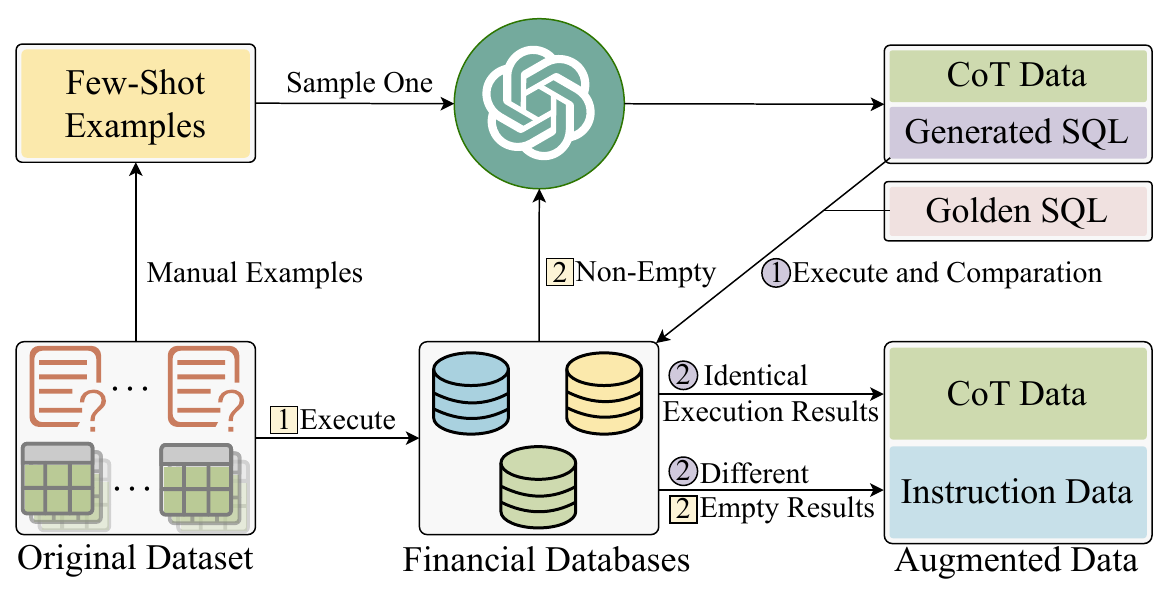}
  \caption{The overview of CoT generation based on self-check}
  \label{fig:CoT}
\end{figure}

To address this issue, we design a specialized prompt template that can guide effectively LLMs in generating correct CoT content. Additionally, we use an execution-based self-check module to filter out incorrect content. Figure~\ref{fig:CoT} provides the overview of the CoT generation progress. 

For a given piece of data, we first place the SQL query into the corresponding database for execution. If the execution yields empty results, we skip that data. Otherwise, we combine it with a one-shot example to create the prompt. Then the prompt is used to instruct the LLMs to generate intermediate reasoning content. Finally, we execute both the SQL generated and the real SQL in the database. If the results of these executions match, we deem the generated data as accurate; otherwise, we discard that data. 

Here, we first filter out the SQL with empty execution results. This step is essential because when both the golden SQL and generated SQL have empty execution results, it becomes challenging to ensure semantic consistency between these two SQL queries. We also perform a self-check step to exclude the generated SQL queries that have different execution results from the golden SQL query. The success rate comparison of generating CoT with and without self-check can be seen in Table~\ref{tab:CoT_success}.

\begin{figure}[tbp]
  \includegraphics[width=\linewidth]{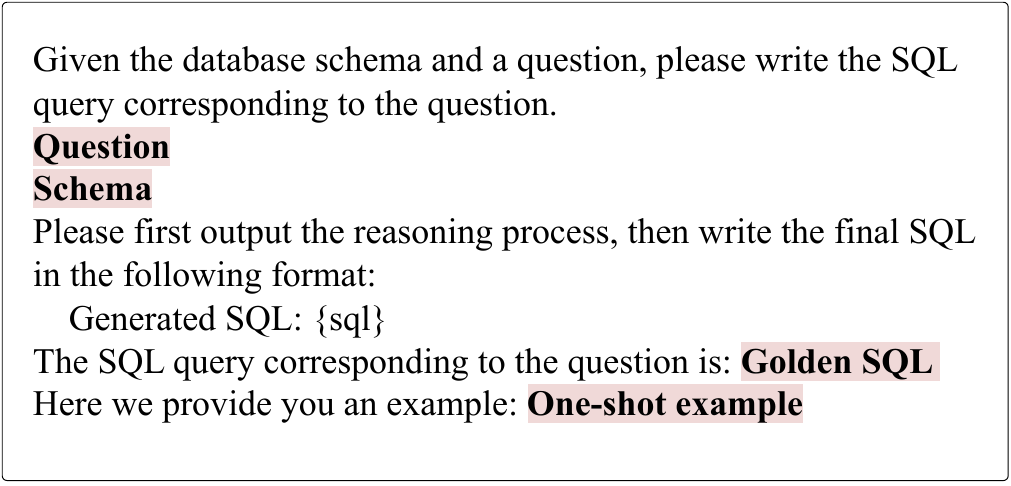}
  \caption{The CoT prompt template. The words in red are the input variables. Here we need to provide the question, schema information, golden SQL and one-shot example to fill the template.}
  \label{fig:CoT_prompt}
\end{figure}

The prompt template used to instruct the LLM can be seen in Figure~\ref{fig:CoT_prompt}. Within this prompt template, we first give the LLM one example of generating CoT content to specify the desired generation style. Then we provide the golden SQL query and ask the LLM to generate CoT content related to it. Since only the CoT content is required rather than the SQL corresponding to the question, we can get excellent CoT content in this way.

\begin{table}[htbp]
\setlength{\tabcolsep}{2.6mm}
  \caption{Success rate of generating CoT with different methods in Chinese dataset.}
  \label{tab:CoT_success}
  \begin{tabular}{cccc}
    \toprule
    Method&Success&Failure&Empty Execution\\
    \midrule
    w self-check & 69.14\% & 18.25\%  & 12.61\%\\
    w/o self-check & 29.95\% & 57.44\%  & 12.61\%\\
  \bottomrule
\end{tabular}
\end{table}

\subsubsection{Synonymous Question Augmentation}
A SQL query can correspond to various natural language questions; however,  in the labeling process for text-to-SQL, the 
annotators typically label a SQL query with only one natural language question. Therefore, the training data cannot well support users' distinct linguistic styles, and it is necessary to expand natural language questions to enrich the diversity of question styles.

\begin{figure}[tbp]
  \includegraphics[width=\linewidth]{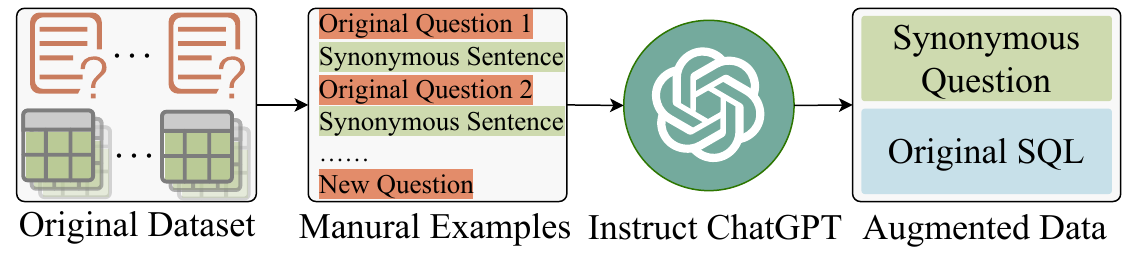}
  \caption{The overview of synonymous question generation}
  \label{fig:synonymous}
\end{figure}

To enrich the diversity of question styles, we utilize ChatGPT to automatically generate synonymous questions, as shown in Figure~\ref{fig:synonymous}. 
We manually write several examples of questions paired with their synonymous sentences as few-shot examples. Subsequently, we concatenate the provided question to these examples to form a comprehensive prompt. This prompt instructs ChatGPT to generate synonymous questions.

\subsubsection{Rule-based Augmentation}
\begin{figure}[tbp]
  \includegraphics[width=\linewidth]{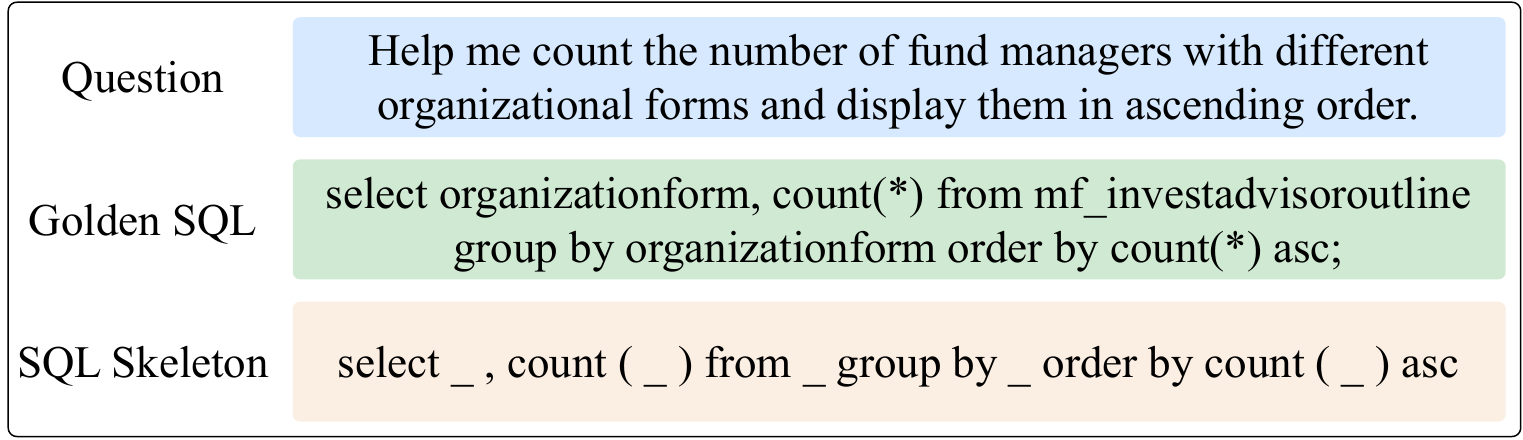}
  \caption{An example of SQL skeleton}
  \label{fig:skeleton}
\end{figure}

In addition to enhancing data quality from the perspective of  natural language questions, we can also enhance data from the perspective of the skeleton/structure of SQL queries. The SQL skeleton contains all the SQL keywords with placeholders for missing identifiers, such as table names and column names. One example of the SQL skeleton can be seen in Figure~\ref{fig:skeleton}. 

\begin{figure}[tbp]
  \includegraphics[width=\linewidth]{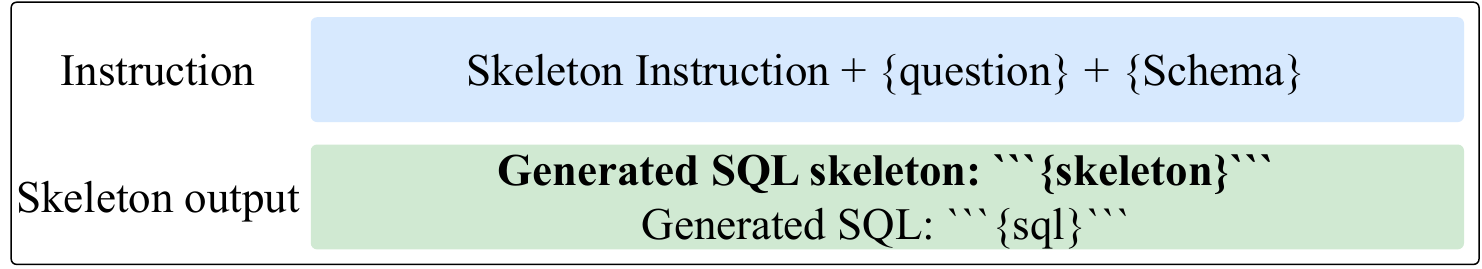}
  \caption{An example of skeleton augmented data}
  \label{fig:skeleton_data}
\end{figure}

We design rules for extracting keywords from SQL queries to obtain their corresponding skeletons. Subsequently, we create the skeleton augmented dataset, as shown in Figure~\ref{fig:skeleton_data}, which instructs the model to generate SQL skeletons first and then generate the final SQL queries during the training stage.

\subsection{Parallel Schema Linking}
Schema Linking, which aims to build connections between question and schema items, is important for improving the performance of Text-to-SQL~\cite{lei2020re}. However, existing state-of-the-art schema linking methods cannot be directly adopted in real-world financial scenarios. Specifically,
Graph neural network (GNN)-based approaches~\cite{graphix-t5} are not readily adaptable to widely-used open-source LLMs such as LLaMA, Baichuan, etc; besides, although Cross-Encoder-based methods~\cite{RESDSQL} produce excellent results and can be developed as a separate module independent of LLMs, they are constrained by the context length limitations of BERT-like models and are not well-suited for handling scenarios involving multiple wide tables.

\begin{figure}[tbp]
  \includegraphics[width=\linewidth]{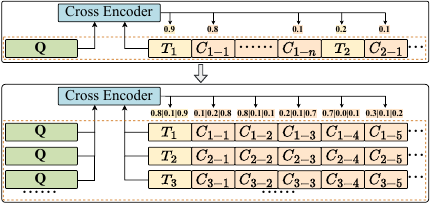}
  \caption{The inference process of Cross-Encoder model}
  \label{fig:Cross-Encoder}
\end{figure}

To effectively perform schema Linking for financial text-to-SQL, we directly utilize the Cross-Encoder model architecture \cite{RESDSQL} and improve the model's training and inference procedure to adapt to the financial scenarios, as shown in Figure~\ref{fig:Cross-Encoder}. 

The original Cross-Encoder model concatenates all tables and columns and then predicts probability values for each of them in a serial manner. As the number of tables and columns grows, the time required for sequential probability prediction increases linearly,
leading to inefficient and slow system performance. Besides, 
when the number of tables and columns is large, the model cannot load all the schema elements and ultimately fails to complete the text-to-SQL task. 

To overcome these issues, we propose a parallel Cross-Encoder model for rapid and accurate retrieval of schema items. Rather than serializing schema items into a single sequence, we organize the tables into a batch, where each element represents a table along with its corresponding column descriptions. The batch, with a size equivalent to the number of tables, is then fed into the modified Cross-Encoder model, enabling it to predict probability values for each table and its associated columns in parallel. This parallel Cross-Encoder model demonstrates the capability to rapidly and accurately retrieve relevant tables and columns from hundreds of schema items. 

\section{Parameter-Efficient Fine-tuning}
Full-parameter fine-tuning, which updates all internal model parameters to handle downstream tasks, faces the following issues:
(1) High computational cost. It usually takes extensive training time to fine-tune an LLM across a substantial number of GPUs. 
(2) High storage cost. After fine-tuning LLMs on downstream tasks, the whole new model parameters are required to be saved, resulting in model weight files of several tens of GBs. 
(3) Lack of cross-database generalization ability. The full-parameter fine-tuning is typically database-specific, which makes the tuned model difficult to generalize to other databases.

To address these issues, this paper proposes a LoRA-based Parameter Efficient Fine-Tuning framework, which supports low-resource fine-tuning and cross-database generalization. This framework consists of a LoRA-based Parameter-Efficient Fine-Tuning (PEFT) method,  a LoRA plugin Hub, and a weights merging method, which enables low computational cost, low storage cost, and cross-database generalization respectively.

\subsection{LoRA-based Multi-Task PEFT}
In this section, we propose a multi-task parameter-efficient fine-tuning method based on Low-Rank Adaptation (LoRA)~\cite{LoRA}. LoRA only tunes the weights of additional rank decomposition matrices and can achieve comparable performance with full-parameter fine-tuning, which can significantly reduce the computational resources required for fine-tuning large models.
Specifically, for a pre-trained weight matrix $W_0 \in R^{d\times k}$, LoRA adds two decomposition matrices on it, denoted as $A$ and $B$, where $A\in R^{d\times r}, B\in R^{r\times k}$ and $r\ll min(d,k)$. Usually, $A$ is initialized using a Gaussian distribution, while $B$ is initialized with zeros. Then the forward process of the pre-trained weight can be represented like this: 
\begin{equation}
    h=W_0x+\Delta Wx=W_0x+BAx
\end{equation}
During the training process, we freeze the original weights of the base model and solely update the weights of the two low-rank matrices, $A$ and $B$.

We perform LoRA on three SQL-related instruction tasks, chain-of-thought generation, synonymous question-driven SQL generation, and skeleton-aware generation. 
Chain-of-thought (CoT) generation task asks the LLMs to output the reasoning process first and then output the final SQL query, which helps to enhance the reasoning ability of LLMs. 
Synonymous question-driven SQL generation replaces the original question with synonymous questions from ChatGPT to instruct the LLMs to generate SQL queries directly, which enables the LLMs to learn meaningful expressions. 
Skeleton-aware generation requires the LLMs to first generate the SQL skeleton and then output the SQL query, leading LLMs to a more comprehensive understanding of SQL structures. 
In these datasets, each table and column concatenates its descriptions, ensuring a comprehensive understanding of the database structure.
We combine these datasets by uniformly mixing them together, which are subsequently used to fine-tune the LLMs using the LoRA method.

\begin{figure}[tbp]
  \includegraphics[width=\linewidth]{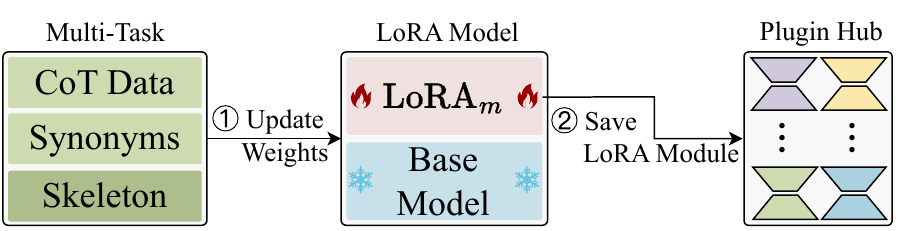}
  \vspace{-0.4cm}
  \caption{The process of fine-tuning LLMs via LoRA.}
  \label{fig:LoRA_training}
  \vspace{-0.3cm}
\end{figure}

Figure~\ref{fig:LoRA_training} gives an overview of the proposed LoRA-based multi-task PEFT method. We use datasets for multiple tasks based on data augmentation presented in Section~\ref{section data augmentation} and select an open-source LLM as the base model. We freeze the weights of this base model, as indicated by the blue portion in the figure. Subsequently, we plugin a LoRA module, as highlighted in red, and continuously update the parameters of LoRA module using the multi-task dataset. Finally, the trained LoRA module is saved to the Plugin Hub which is introduced in Section~\ref{sec:lora_plugin}.

\subsection{LoRA Plugin Hub} 
\label{sec:lora_plugin}
By using the LoRA-based multi-task PEFT method, we can efficiently obtain LoRA modules. Furthermore, the LoRA modules are independent of the base model and have a small size (typically less than 100 MB). Therefore, we can train a series of LoRA modules for different uses. For example, we can train a LoRA module for each database to perform database-specific text-to-SQL. We can also train a LoRA module on all the given databases jointly to obtain cross-database generalization ability. This paper stores these LoRA modules together and builds a LoRA module hub, which is dubbed as LoRA plugin hub for the LoRA modules are independent of the base model and can be regarded as plugins for the base model. Besides, the LoRA plugin hub can include LoRA plugins trained on different base models, such as LLaMA2~\cite{llama2}, Baichuan2~\cite{baichuan2}. The LoRA plugin hub is able to support few-shot fine-tuning for low-resource scenarios, which is introduced in Section~\ref{sec:Weights_Mergin}.

\begin{figure}[tbp]
  \includegraphics[width=\linewidth]{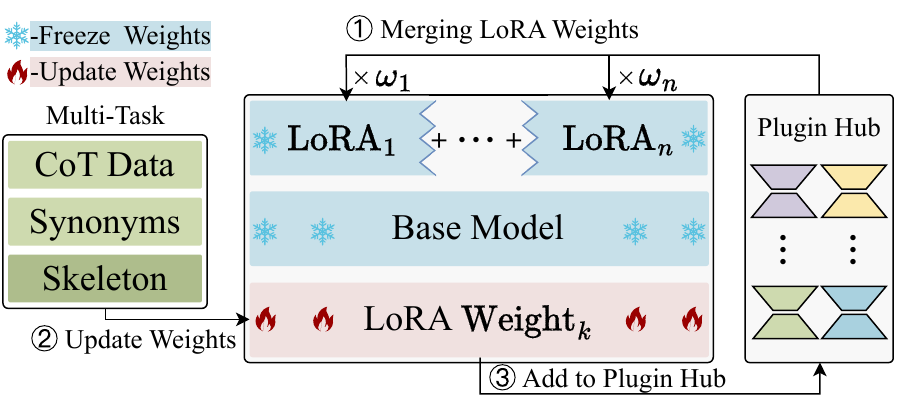}
  \vspace{-0.4cm}
  \caption{The process of few-shot LoRA-based fine-tuning with weights merging.}
  \label{fig:few_shot_training}
  \vspace{-0.2cm}
\end{figure}

\subsection{Weights Merging-based Few-shot LoRA}
\label{sec:Weights_Mergin}
In practice, it is common that new databases are rapidly constructed with business growth. For a new database, the number of training data is limited; thus, it is necessary to perform few-shot fine-tuning to overcome such a low-resource scenario. This paper proposes a weights merging-based few-shot LoRA-based fine-tuning method to achieve efficient cross-database generalization and handle the text-to-SQL tasks for new databases.

In this paper, the weights merging is conducted over the LoRA plugins stored in the LoRA plugin hub. Specifically, we take several related LoRA plugins from the LoRA Plugin Hub  and  merge them into a single LoRA plugin by means of weighted summation. The merged LoRA plugin can be represented in the following formula:

\begin{equation}
    \hat{A} = \omega_1 A_1 + \omega_2 A_2 + \cdots + \omega_n A_n 
\end{equation}
\begin{equation}
    \hat{B} = \omega_1 B_1 + \omega_2 B_2 + \cdots + \omega_n B_n
\end{equation}
where $A_i, B_i$ $(i \in [1, n])$ are the low-rank matrices of the $i$-th LoRA plugin, $\hat{A}, \hat{B}$ are the low-rank matrices of the new LoRA plugin, $\omega_i$ is the weight coefficient of each LoRA plugin and $n$ is the number of LoRA plugins to be merged. 

After that, we initialize the base model with the merged LoRA plugin. Subsequently, we plug an additional LoRA plugin, denoted as $A_k, B_k$ into the base model and further fine-tune it on this foundation. 
Then the forward process of a pre-trained weight of the base model can be represented as: 
\begin{equation}
    \begin{aligned}
        \hat{h} &= W_0x+\Delta \hat{W}x + \Delta W_k x \\
                &= W_0x+\hat{B}\hat{A}x + B_k A_kx \\ 
                &= W_0x+(\sum_{i=1}^n\omega_i A_i)(\sum_{i=1}^n\omega_i B_i)x + B_k A_kx.
    \end{aligned}
\end{equation}
where $A_k$ and $B_k$ represent the additional LoRA weights that require updates through fine-tuning. This approach allows LLMs to leverage the knowledge and ability encoded in trained LoRA plugins from various domains, enhancing their performance on the target domain database. 

Figure~\ref{fig:few_shot_training} illustrates the process of the weights merging-based few-shot LoRA-based fine-tuning method. It first extracts LoRA modules from Plugin Hub and merges them. Then, the multi-task dataset generated from Section~\ref{section data augmentation} is used for fine-tuning the LoRA model. After fine-tuning, the trained LoRA module is saved to the Plugin Hub.

\section{Output Calibration}

\begin{figure}[tbp]
  \includegraphics[width=\linewidth]{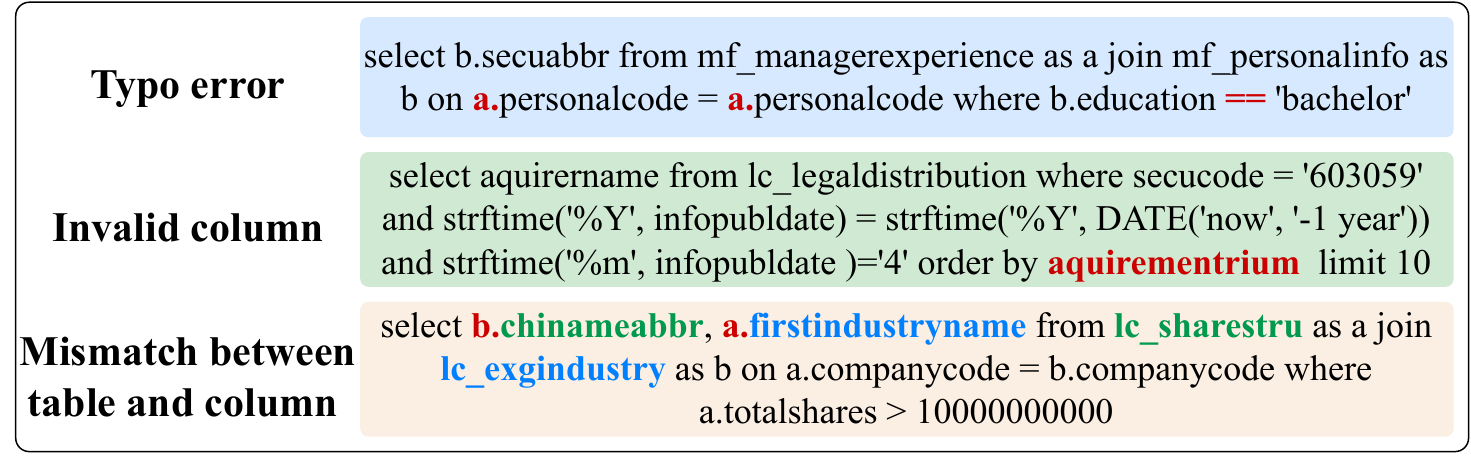}
  \caption{The examples of invalid SQL queries generated by LLMs. The first SQL makes syntactical mistakes in join and where conditions. The second one uses an invalid column, as column \textit{aquirementrium} is nonexistent in the database. The true column in the golden SQL is \textit{aquireramount}. The third SQL builds the wrong connections between tables and columns. The column \textit{chinameabbr} and \textit{firstindustryname} belong to table \textit{lc\_sharestru} and \textit{lc\_exgindustry} respectively.}
  \label{fig:output_example}
\end{figure}

Due to the hallucination and decoding strategy of LLMs, LLM-based text-to-SQL models often generate invalid and syntactically incorrect SQL queries. For instance, there may be non-existent table and column names, and incorrectly associated relationships between tables and columns in the generated SQL queries. There are some examples of the generated invalid SQL queries, as shown in Figure~\ref{fig:output_example}.

\begin{algorithm}[tbp]
\caption{Output Calibration Step}
\KwIn{the list of $n$ candidate SQL queries $Q$; \\ the schema information of the SQL queries $S$}
\KwOut{the final SQL query $\hat{q}$}
\BlankLine
\emph{initialize $E=Dict\{\}$, which maps SQL to SQL keyword components}\;
\ForEach{$q_i \in Q $}{
    \state{$q_i \leftarrow f_1(q_i, S)$ \ \ \tcp*[h]{fix some typo errors}}\;
    \state{$e_i \leftarrow f_2(q_i, S)$ \ \ \tcp*[h]{extract keyword components}}\;
    \If{columns of $e_i$ in $S$ }{ 
        $E[q_i]=e_i$
    }
}
\emph{initialize $C=List()$, the cluster list of the SQL queries}\;
\ForEach{$q_i, e_i \in E$}{
    \ForEach{$C_j \in C$}{
        \state{$q_j=C_j[0], e_j=E[q_j]$} \\
        \If{$e_i$ compatible with $e_j$ }{
            \state{$C_j$.append($q_i$)} \\
            \state{break}
        }
    }
    \If{$q_i$ doesn't match any cluster in $C$}{
        $C$.append($[q_i]$)
    }
}
\emph{sort all $C_i \in C$ in descending order according to $len(C_i)$}\;
\state{$\hat{q} \leftarrow C_1[0]$} \\
\state{$\hat{q} \leftarrow f_3(\hat{q}, S)$ \ \ \tcp*[h]{align tables to columns}}

return $\hat{q}$
\label{alg:alg1}
\end{algorithm}

To calibrate the output of LLMs and enhance its correctness, this paper proposes an efficient output calibration algorithm. The proposed method calibrates the outputs without executing the SQL queries, which addresses the execution-dependency issue that occurred in the state-of-the-art methods (e.g., NatSQL~\cite{natsql, RESDSQL}, execution-based self-consistency~\cite{sqlpalm, dong2023c3}) and can be directly used in the real financial applications. 
The details of this method are presented in  Algorithm~\ref{alg:alg1}.

The input of the algorithm comprises a list of $n$ candidate SQL queries generated in parallel by the LLMs, along with the corresponding schema information for these SQL queries. The objective of our algorithm is to identify the most consistent and valid SQL query.

Due to the inherent randomness of LLMs, the SQL queries they generate frequently contain grammatical errors, which hinders the process of SQL parsing. Thus, we rectify some typo errors for each SQL query, such as replacing " == " with " = " or addressing issues where the "JOIN ON" keyword is used without specifying the corresponding foreign key. 
When encountering an invalid column, we employ a fuzzy matching approach to replace it with the column from the schema that is most similar in terms of characters. 
These problems would affect the subsequent extraction of SQL keywords and their values. Subsequently, we parse the modified SQL and extract the keywords with their 
corresponding values, which are used to apply a non-execution-based self-consistency method. Specifically, we determine the equivalence of two SQL queries by assessing the consistency of their SQL keywords and values. Using this criterion, we cluster equivalent SQL queries together. Then, we select one SQL query from the largest cluster as the result. This approach significantly enhances the consistency of SQL queries.

Finally, we verify whether the SQL query associates tables with columns and then align the tables with their respective columns. For every column specified in the SQL, this module guarantees that the corresponding table must be included in the $FROM$ clause.
In instances where $table.column$ or $alias.column$ is encountered, the module verifies whether the specified $column$ is indeed associated with the specified $table$ or the $alias\ table$. 
If not, the module searches for the appropriate $table$ containing the specified $column$ within the $FROM$ clause. If the table is not found through this search, the module selects one table from the schema information. 

Traditional SQL calibrating methods involve executing the SQL queries within the databases to identify and rectify errors, which enhances the overall quality of the SQL. However, such approaches are not applicable to financial analysis scenarios, where the databases are often huge, and executing SQL queries can be time-consuming. In contrast, our output calibration algorithm relies on the inherent structure and syntactical rules of SQL, efficiently mitigating the impact of illusions and uncertainty within LLMs when calibrating SQL queries. Through such calibration of SQL queries, our approach significantly improves the consistency and utility of the final SQL queries.

\section{Experiments}
In this section, we perform experimental studies to validate the effectiveness of FinSQL in financial Text-to-SQL. We first compare the overall performance of GPT-based methods and T5-based methods. Subsequently, we display the effectiveness of our parallel Cross-Encoder model. Then we perform ablation studies on our Hybrid Data Augmentation and Output Calibration methods. Additionally, we conduct substantial experiments to validate the excellent performance of our weights merging methods in few-shot cross-database model transfer. 

\subsection{Experiment Setup}

\textbf{Datasets}. 
We conduct experiments on our financial Text-to-SQL dataset BULL, which is collected from real industrial scenarios. This dataset comprises three databases, across three common financial domains: fund, stock, and macro economy. On average, each database contains 26 tables and 390 columns. It includes 4966 question-SQL query pairs, with 1744 training examples and 405 development examples for fund, 1672 training examples and 464 development examples for stock, and 550 training examples and 131 development examples for macro economy. Notably, The dataset is available in both English (denoted as BULL-en) and Chinese (denoted as BULL-cn) versions. 

\textbf{Evaluation Metrics}.
We choose execution accuracy (EX) as our evaluation metric, as implemented by Test Suite Accuracy~\cite{test_suite}. This metric is also the official evaluation metric used by the popular Text-to-SQL leaderboard, Spider~\cite{spider}. EX executes the predicted SQL query and golden SQL query in the database and judges whether the two have the same execution results. 

\textbf{Implementation}.
We employ both decoder-only architecture models, LLaMA2~\cite{llama2} and Baichuan2~\cite{baichuan2}, as well as encoder-decoder architecture models, T5~\cite{t5} and mT5~\cite{mt5}, as the base large language models.
LLaMA2 comprises a range of large language models with a parameter scale spanning from 7B to 70B. LLaMA is currently one of the most effective and influential open-source large language models. It holds a leading position in many English benchmarks compared to other open-source models. Therefore, we choose LLaMA2 as the base model and fine-tune it on BULL-en. 
Baichuan2 is a collection of large multilingual language models, encompassing 7B and 13B parameters, trained from scratch on a corpus of 2.6 trillion tokens. Baichuan2 demonstrates remarkable performance on well-established benchmarks. Considering its proficiency in the Chinese language, we fine-tune it on BULL-cn. T5, an encoder-decoder model,  is pre-trained on an amount of unsupervised and supervised multi-tasks. Additionally, mT5 is a multilingual variant of T5 that covers 101 languages. Methods based on T5 achieve SOTA on the Sider leaderboard of fine-tuning setting. Therefore, we also choose T5 and mT5 as the base models to handle English and Chinese tasks, respectively. 
We employ Cross-Encoder models based on Roberta-large~\cite{roberta} and Chinese-Bert-large~\cite{chinese_roberta} for BULL-en and BULL-cn, respectively. 
All the fine-tuning experiments are conducted with LoRA on a single A40 GPU. 

\textbf{Baseline}.
We conduct experiments on our financial Dataset and compare it with the following baselines: 
(1) \textbf{DAIL-SQL}~\cite{dail-sql} improves the selection process by encoding structured knowledge as SQL statements, selecting examples based on skeleton similarities, and removing cross-domain knowledge from examples for token efficiency. 
(2) \textbf{DIN-SQL}~\cite{DIN-SQL} enhances the performance of LLM-based text-to-SQL models by implementing a strategic task decomposition approach. It also brought in adaptive prompting techniques that are uniquely adjusted to the complexity of specific tasks.
(3) \textbf{C3}~\cite{dong2023c3} integrates Clear Prompting, Calibration with Hints, and Consistent Output, significantly enhancing accuracy and systematizing processes in GPT-based Text-to-SQL tasks.
(4) \textbf{RESDSQL}~\cite{RESDSQL} proposes a ranking-enhanced encoding and skeleton-aware decoding framework, which utilizes a two-stage method to retrieve relevant schema items first and then generate the SQL queries based on T5. It is the best fine-tuning based method on Spider leaderboard; 
(5) \textbf{Token Preprocessing}~\cite{Token_Preprocessing} inserts spaces to separate words within schema and question  tokens, improving their readability and semantic clarity. 
(6) \textbf{Picard}~\cite{picard} employs incremental parsing to restrict auto-regressive decoders in language models, effectively filtering out unsuitable tokens to improve the precision of text-to-SQL translations.

\subsection{Overall Performance}
In Table~\ref{tab:overall results en} and Table~\ref{tab:overall results zh}, we report the performance of our FinSQL method against other leading baseline methods from Spider leaderboard on our proposed dataset of English and Chinese versions respectively. 
On the BULL-en, we combine FinSQL with LLaMA2 and T5 to compare with GPT-based methods and T5-based methods respectively. On the BULL-cn, we employ Baichuan2 and mT5 to better adapt to the Chinese context. Our FinSQL method outperforms all other approaches on both BULL-en and BULL-cn. On BULL-en, FinSQL exhibits an EX of 82.2\% with LLaMA and 81.5\% with T5. On BULL-cn, FinSQL displays an EX of 76.6\% with Baichuan2 and 70.4\% with mT5.

\begin{table}[tbp]
\setlength{\tabcolsep}{2.1mm}
  \caption{Overall results of different previous methods on BULL-en. For fine-tuning methods, we employ T5-large and LLaMA2-13B. The * means we employ our parallel Cross-Encoder model for schema linking.}
\vskip -0.05in 
  \label{tab:overall results en}
  \begin{tabular}{ccc}
    \toprule
    Model&EX & Cost Per SQL(\$)\\
    \midrule
    DIN-SQL~\cite{DIN-SQL} + GPT-4 & - & 4.9103\\
    DAIL-SQL~\cite{dail-sql} + GPT-4 & 75.0 & 0.1579 \\
    DAIL-SQL~\cite{dail-sql} + ChatGPT & 46.0 & 0.0051 \\
    C3\cite{dong2023c3} + ChatGPT~ & 7.0 & 0.0065\\
    \midrule
    RESDSQL*~\cite{RESDSQL} + T5 & 78.8 & - \\
    Token Preprocessing*~\cite{Token_Preprocessing} + T5 & 67.5 & - \\
    Picard*~\cite{picard} + T5 & 79.3 & - \\
    \midrule
    FinSQL + LLaMA2 &  82.2 & - \\
    FinSQL + T5& 81.5 & -\\
  \bottomrule
\end{tabular}
\end{table}

\begin{table}[tbp]
\setlength{\tabcolsep}{1.75mm}
  \caption{Overall results of different previous methods on our BULL-cn. For fine-tuning methods, we employ mT5-large and Baichuan2-13B. The * means we employ our parallel Cross-Encoder model for schema linking.}
\vskip -0.05in 
  
  \label{tab:overall results zh}
  \begin{tabular}{ccc}
    \toprule
    Model&EX & Cost Per SQL(\$)\\
    \midrule
    DIN-SQL~\cite{DIN-SQL} + GPT-4 & - & 4.9158\\
    DAIL-SQL~\cite{dail-sql} + GPT-4 & 75.0 & 0.1581 \\
    DAIL-SQL~\cite{dail-sql}+ ChatGPT & 55.0 & 0.0053\\
    C3~\cite{dong2023c3} + ChatGPT & 2.0 & 0.0078\\
    \midrule
    RESDSQL*~\cite{RESDSQL} +  mT5 & 66.9 & - \\
    Token Preprocessing*~\cite{Token_Preprocessing} + mT5 & 60.2 & - \\
    Picard*~\cite{picard} + mT5 & 72.7 & - \\
    \midrule
    FinSQL + Baichuan2& 76.6  & -\\
    FinSQL + mT5 & 70.4 & - \\ 
  \bottomrule
\end{tabular}
\vskip -0.1in 
\end{table}

\textbf{GPT-based methods}. 
For methods utilizing GPT-4 and ChatGPT, we calculate the Cost Per SQL based on the unit price in Table~\ref{tab:price} and the number of input and output tokens. Due to budget constraints, we only select 20 entries for GPT-4 based methods and 100 entries for ChatGPT based methods. For DIN-SQL + GPT-4, the total length of the prompt exceeds GPT-4's token limit of 8192, rendering the test unfeasible. Although GPT-4-32k can meet our needs of context length, the price is too high for us to afford, hence we only estimate the api cost. 
The methods based on GPT show varied results: DAIL-SQL + GPT-4 achieves a considerable EX of 75\%, which is around 7\% lower than FinSQL + LLaMA2 and 1.6\% lower than FinSQL + Baichuan2. On BULL-en, The EX of DAIL-SQL + ChatGPT and C3 + ChatGPT is 46\% and 7\%  respectively. On BULL-cn, the numbers are 55\% and 2\%, which are not comparable with the other methods. We speculate that this is due to their specific prompts for Spider dataset, which results in exhibiting poor performance when facing data variability. In addition to incurring extra costs, GPT-based methods require several hours to infer results from less than 100 data entries. In contrast, FinSQL is better suited to adapt to real-world data scenarios, thus holding high practical value and cost-effectiveness. 

\textbf{T5-based methods}. 
Due to context limitations, T5-based methods are unable to process our concatenated input sequences. Therefore, we initially utilize our parallel Cross-Encoder for retrieval, simplifying each question's schema to 3 tables and 7 columns. As a result, T5-based methods achieve a higher accuracy compared to GPT-based methods. they are all able to achieve an execution accuracy rate of over 60\%. It is worth noting that Picard demonstrates a substantial accuracy rate, making it the best-performing method after FinSQL. We surmise that this is due to its additional filtering of irrelevant tokens. Besides, as for methods applied to mT5, FinSQL performs 2.3\% lower than Picard. Although FinSQL still has shortcomings in BULL-ch, it has provided valuable insights into the challenges and complexities of Text-to-SQL in financial analysis, guiding our future research directions.

\begin{table}[tbp]
\setlength{\tabcolsep}{7.1mm}
  \caption{Performance of Schema Linking}
  \label{tab: schema linking performance}
  \begin{tabular}{ccc}
    \toprule
    Schema Item&Table&Column \\
    \midrule
    AUC (BULL-en) & 0.9984 & 0.9979 \\
    AUC (BULL-cn) & 0.9995 & 0.9994 \\ 
  \bottomrule
\end{tabular}
\end{table}

\begin{table}[tbp]
\setlength{\tabcolsep}{1.82mm}
\caption{The recall@k of tables and columns of our Parallel Cross-Encoder model.}
\centering
\begin{tabular}{lccccccc}
\hline
 \multirow{2}*{\textbf{Dataset}} & \multicolumn{3}{c}{\textbf{Table}} & & \multicolumn{3}{c}{\textbf{Column}} \\
\cline{2-4} 
\cline{6-8}
 & R@3 & R@5 & R@10 & & R@5 & R@7 & R@10 \\
\hline
BULL-en & 99.1 & 99.5 & 99.8&  & 96.6 & 98.0 & 99.0 \\
BULL-cn & 99.5 & 99.9 & 100.0& & 97.1 & 98.6 & 99.5 \\
\hline
\end{tabular}
\label{tab:my_label}
\end{table}

\begin{table}[tbp]
\setlength{\tabcolsep}{2.2mm}
  \caption{Effective of different data augmentation methods. }
  \label{tab: ablation study augmentation}
  \begin{tabular}{ccc}
    \toprule
    Technique &EX (English) &EX (Chinese)\\
    \midrule
    Hybrid Data Augmentation  & 80.7 &  75.3\\ 
    \midrule
    w/o CoT Data  & 78.3 (-2.4) & 73.4 (-1.9)\\
    w/o Synonyms Data  & 77.0 (-2.7) & 73.8 (-1.5)\\
    w/o Skeleton Data  & 78.1 (-2.6) & 71.0 (-4.3)\\ 
    w/o Augmented Data  & 76.9 (-3.8) & 70.6 (-4.7)\\ 
  \bottomrule
\end{tabular}
\vskip -0.1in
\end{table}

\begin{figure*}[tbp]
  \centering
    \vskip -0.1in
  \includegraphics[width=\textwidth]{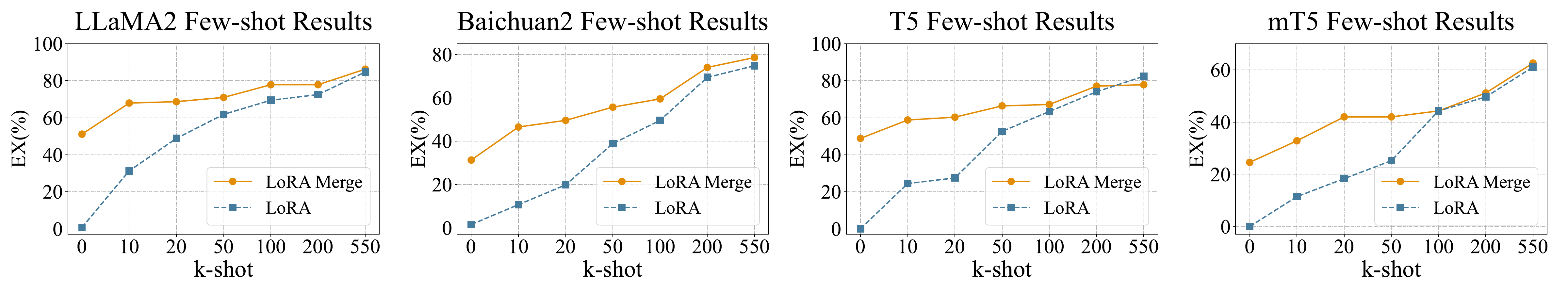}
  \caption{Execution accuracy of weights merging-based few-shot
LoRA on four models.}
  \label{fig:few-shot}
\end{figure*}

\subsection{Effect of Schema Linking}
To evaluate the efficacy of our schema linking method, we adopt the Area Under the ROC Curve (AUC) as our metric. As presented in Table~\ref{tab: schema linking performance}, the outcomes reveal an AUC of 0.9984 for tables and 0.9979 for columns in the BULL-en, alongside 0.9995 for tables and 0.9994 for columns in the BULL-cn. These metrics validate the robustness of our approach, demonstrating the precise identification of relevant schema items in both datasets.

Additionally, we conduct experiments on our parallel cross-encoder model, with detailed results in Table~\ref{tab:my_label}. The model's effectiveness is measured using the recall at k (R@k) metric, which evaluates its ability to correctly identify relevant items within the top k selections. In the BULL-en, the model exhibite outstanding results for table recall, achieving 99.1\%, 99.5\%, and 99.8\% for R@3, R@5, and R@10 respectively. In BULL-cn, it shows superior results with scores of 99.5\%, 99.9\%, and a perfect 100.0\%. In the more complex task of column recall, the model sustains strong performance with 96.6\%, 98.0\%, and 99.0\% for R@5, R@7, and R@10 in BULL-en, and 97.1\%, 98.6\%, and 99.5\% in BULL-cn. These results highlight the model's proficiency in managing multilingual datasets with remarkable precision and recall, a crucial aspect for the subsequent stages of Text-to-SQL task.

\subsection{Effect of Data Augmentation}
In this section, we perform an ablation study to examine the impact of various data augmentation techniques. 
To eliminate influence from other unrelated factors and focus solely on the impact of data augmentation, we did not perform output calibration in this study. The experimental results are illustrated in Table~\ref{tab: ablation study augmentation}. It is evident that the exclusion of each specific augmentation method degrades the performance. When CoT Data is omitted, the execution accuracy experiences a reduction of 1.9\% for Chinese and 2.4\% for English, confirming the significance of this particular data enhancement. Similarly, the absence of Synonyms Data leads to a decrease of 1.5\% for Chinese and 2.7\% for English, while the removal of Skeleton Data results in a decline of 4.3\% for Chinese and 2.6\% for English. The most notable impact is observed when all forms of Augmented Data are excluded, culminating in a decrease of 4.7\% for Chinese and 3.8\% for English. These results collectively validate the necessity of the hybrid data augmentation approach in preserving the robustness of FinSQL.

\subsection{Effect of Weights Merging-based Few-shot LoRA}
In this section, we discuss the role of our Weights Merging-based Few-shot LoRA in a low-sample scenario. In our dataset, there are 1744 and 1672 training samples for fund and stock respectively, while macro economy only has 550 data samples. Consequently, we separately train LoRA models on fund and stock datasets, and subsequently merge these two LoRA modules using the average weighting. Following the merging process, we continue fine-tuning the merged LoRA model with varying amounts of data in macro economy. We also train the Cross-Encoder model with the combination data of funds, stocks, and varying amounts of macro data to retrieve relevant schema items. The results are shown in Figure~\ref{fig:few-shot}.

From the Figure, we can observe that whether it is LLaMA2, Baichuan2, T5, or mT5, the performance using LoRA Merge is usually higher than LoRA, with a greater disparity in performance at lower k-shot values. Across the four models, for the zero-shot learning setting, the performance of LoRA-Merge surpasses that of LoRA by 50.39\% in LLaMA2, 29.77\% in Baichuan2, 47.33\% in T5, and 24.57\% in mT5. For 10-shot learning, LoRA-Merge's performance exceeds LoRA by 36.64\%, 35.87\%, 34.35\%, and 21.37\% respectively. As the number of few-shot instances increases, the disparity gradually diminishes. When the shot number exceeds 200, the trained LoRA sometimes performs better than the merged LoRA. However, we can still observe the significant benefits of merging LoRA weights. When there is a scarcity of training data, the pre-merged LoRA module fine-tuned on other tasks exhibits a more pronounced performance than LoRA module trained from scratch. Hence this method is particularly suitable for low-sample scenarios.

\begin{table}[tbp]
\setlength{\tabcolsep}{2.98mm}
  \caption{Effect of Output Calibration }
  \label{tab: ablation study}
  \begin{tabular}{ccc}
    \toprule
    Technique&EX (Chinese) &EX (English)\\
    \midrule
    FinSQL &  76.6 & 82.2\\ 
    \midrule
    w/o Output Calibration & 75.3 (-1.3) & 80.7 (-1.5) \\
    w/o Self-Consistency & 76.4 (-0.2) & 81.9 (-0.3) \\
    w/o Alignment & 75.5 (-1.1) & 81.0 (-1.2)\\
  \bottomrule
\end{tabular}
\end{table}

\subsection{Effect of Output Calibration}
To investigate the impact of output calibration techniques on model performance, we conduct an ablation study as depicted in Table~\ref{tab: ablation study}. The output calibration is comprised of two methodologies: self-consistency and alignment. The results are quantified in terms of execution accuracy for Chinese and English.

Our findings indicate that the exclusion of output calibration results in a reduction of EX accuracy by 1.3\% for Chinese and 1.5\% for English. The absence of self-consistency leads to a downtrend of 0.2\% for Chinese and 0.3\% for English. Lastly, dropping alignment leads to a decrease of 1.1\% in EX accuracy for Chinese and 1.2\% for English. As expected, the cumulative impact of removing self-consistency and alignment equates to the elimination of output calibration. This further confirms that the observed decrease in EX is indeed related to the elimination of output calibration in both linguistic contexts.
\section{Conclution}
In this paper, we propose BULL, a Text-to-SQL dataset derived from real-world financial scenarios. Based on this dataset, we propose FinSQL, a model-agnostic LLMs-based Text-to-SQL framework for financial analysis. 
FinSQL conducts hybrid data augmentation and parallel Cross-Encoder to construct concise and diversified prompts. 
Besides, FinSQL utilizes LoRA merging methods to improve the performance for cross-database model transfer. 
Moreover, FinSQL proposes an output calibration method to improve the quality of SQL queries generated by LLMs. 
Extensive experiments on BULL demonstrate the effectiveness of FinSQL in financial analysis. 
The BULL and FinSQL have the potential to forge a new trend in financial Text-to-SQL research.

\newpage
\clearpage
\balance

\bibliographystyle{ACM-Reference-Format}
\bibliography{sample-sigconf}


\begin{thebibliography}{41}


\ifx \showCODEN    \undefined \def \showCODEN     #1{\unskip}     \fi
\ifx \showDOI      \undefined \def \showDOI       #1{#1}\fi
\ifx \showISBNx    \undefined \def \showISBNx     #1{\unskip}     \fi
\ifx \showISBNxiii \undefined \def \showISBNxiii  #1{\unskip}     \fi
\ifx \showISSN     \undefined \def \showISSN      #1{\unskip}     \fi
\ifx \showLCCN     \undefined \def \showLCCN      #1{\unskip}     \fi
\ifx \shownote     \undefined \def \shownote      #1{#1}          \fi
\ifx \showarticletitle \undefined \def \showarticletitle #1{#1}   \fi
\ifx \showURL      \undefined \def \showURL       {\relax}        \fi
\providecommand\bibfield[2]{#2}
\providecommand\bibinfo[2]{#2}
\providecommand\natexlab[1]{#1}
\providecommand\showeprint[2][]{arXiv:#2}

\bibitem[Cao et~al\mbox{.}(2021)]%
        {lgesql}
\bibfield{author}{\bibinfo{person}{Ruisheng Cao}, \bibinfo{person}{Lu Chen}, \bibinfo{person}{Zhi Chen}, \bibinfo{person}{Yanbin Zhao}, \bibinfo{person}{Su Zhu}, {and} \bibinfo{person}{Kai Yu}.} \bibinfo{year}{2021}\natexlab{}.
\newblock \showarticletitle{{LGESQL:} Line Graph Enhanced Text-to-SQL Model with Mixed Local and Non-Local Relations}. In \bibinfo{booktitle}{\emph{ACL}}.
\newblock


\bibitem[Chowdhery et~al\mbox{.}(2023)]%
        {palm}
\bibfield{author}{\bibinfo{person}{Aakanksha Chowdhery}, \bibinfo{person}{Sharan Narang}, \bibinfo{person}{Jacob Devlin}, \bibinfo{person}{Maarten Bosma}, \bibinfo{person}{Gaurav Mishra}, \bibinfo{person}{Adam Roberts}, \bibinfo{person}{Paul Barham}, \bibinfo{person}{Hyung~Won Chung}, \bibinfo{person}{Charles Sutton}, \bibinfo{person}{Sebastian Gehrmann}, \bibinfo{person}{Parker Schuh}, \bibinfo{person}{Kensen Shi}, \bibinfo{person}{Sasha Tsvyashchenko}, \bibinfo{person}{Joshua Maynez}, \bibinfo{person}{Abhishek Rao}, \bibinfo{person}{Parker Barnes}, \bibinfo{person}{Yi Tay}, \bibinfo{person}{Noam Shazeer}, \bibinfo{person}{Vinodkumar Prabhakaran}, \bibinfo{person}{Emily Reif}, \bibinfo{person}{Nan Du}, \bibinfo{person}{Ben Hutchinson}, \bibinfo{person}{Reiner Pope}, \bibinfo{person}{James Bradbury}, \bibinfo{person}{Jacob Austin}, \bibinfo{person}{Michael Isard}, \bibinfo{person}{Guy Gur{-}Ari}, \bibinfo{person}{Pengcheng Yin}, \bibinfo{person}{Toju Duke}, \bibinfo{person}{Anselm Levskaya},
  \bibinfo{person}{Sanjay Ghemawat}, \bibinfo{person}{Sunipa Dev}, \bibinfo{person}{Henryk Michalewski}, \bibinfo{person}{Xavier Garcia}, \bibinfo{person}{Vedant Misra}, \bibinfo{person}{Kevin Robinson}, \bibinfo{person}{Liam Fedus}, \bibinfo{person}{Denny Zhou}, \bibinfo{person}{Daphne Ippolito}, \bibinfo{person}{David Luan}, \bibinfo{person}{Hyeontaek Lim}, \bibinfo{person}{Barret Zoph}, \bibinfo{person}{Alexander Spiridonov}, \bibinfo{person}{Ryan Sepassi}, \bibinfo{person}{David Dohan}, \bibinfo{person}{Shivani Agrawal}, \bibinfo{person}{Mark Omernick}, \bibinfo{person}{Andrew~M. Dai}, \bibinfo{person}{Thanumalayan~Sankaranarayana Pillai}, \bibinfo{person}{Marie Pellat}, \bibinfo{person}{Aitor Lewkowycz}, \bibinfo{person}{Erica Moreira}, \bibinfo{person}{Rewon Child}, \bibinfo{person}{Oleksandr Polozov}, \bibinfo{person}{Katherine Lee}, \bibinfo{person}{Zongwei Zhou}, \bibinfo{person}{Xuezhi Wang}, \bibinfo{person}{Brennan Saeta}, \bibinfo{person}{Mark Diaz}, \bibinfo{person}{Orhan Firat},
  \bibinfo{person}{Michele Catasta}, \bibinfo{person}{Jason Wei}, \bibinfo{person}{Kathy Meier{-}Hellstern}, \bibinfo{person}{Douglas Eck}, \bibinfo{person}{Jeff Dean}, \bibinfo{person}{Slav Petrov}, {and} \bibinfo{person}{Noah Fiedel}.} \bibinfo{year}{2023}\natexlab{}.
\newblock \showarticletitle{PaLM: Scaling Language Modeling with Pathways}.
\newblock \bibinfo{journal}{\emph{Journal Of Machine Learning Research}}  \bibinfo{volume}{24} (\bibinfo{year}{2023}), \bibinfo{pages}{240:1--240:113}.
\newblock


\bibitem[Cobbe et~al\mbox{.}(2021)]%
        {gsm8k}
\bibfield{author}{\bibinfo{person}{Karl Cobbe}, \bibinfo{person}{Vineet Kosaraju}, \bibinfo{person}{Mohammad Bavarian}, \bibinfo{person}{Mark Chen}, \bibinfo{person}{Heewoo Jun}, \bibinfo{person}{Lukasz Kaiser}, \bibinfo{person}{Matthias Plappert}, \bibinfo{person}{Jerry Tworek}, \bibinfo{person}{Jacob Hilton}, \bibinfo{person}{Reiichiro Nakano}, \bibinfo{person}{Christopher Hesse}, {and} \bibinfo{person}{John Schulman}.} \bibinfo{year}{2021}\natexlab{}.
\newblock \showarticletitle{Training Verifiers to Solve Math Word Problems}.
\newblock \bibinfo{journal}{\emph{arXiv preprint arXiv:2110.14168}} (\bibinfo{year}{2021}).
\newblock


\bibitem[Cui et~al\mbox{.}(2020)]%
        {chinese_roberta}
\bibfield{author}{\bibinfo{person}{Yiming Cui}, \bibinfo{person}{Wanxiang Che}, \bibinfo{person}{Ting Liu}, \bibinfo{person}{Bing Qin}, \bibinfo{person}{Shijin Wang}, {and} \bibinfo{person}{Guoping Hu}.} \bibinfo{year}{2020}\natexlab{}.
\newblock \showarticletitle{Revisiting Pre-Trained Models for {C}hinese Natural Language Processing}. In \bibinfo{booktitle}{\emph{EMNLP Findings}}.
\newblock


\bibitem[Dai et~al\mbox{.}(2023)]%
        {2023arXiv230213007D}
\bibfield{author}{\bibinfo{person}{H Dai}, \bibinfo{person}{Z Liu}, \bibinfo{person}{W Liao}, \bibinfo{person}{X Huang}, \bibinfo{person}{Y Cao}, \bibinfo{person}{Z Wu}, \bibinfo{person}{L Zhao}, \bibinfo{person}{S Xu}, \bibinfo{person}{W Liu}, \bibinfo{person}{N Liu}, {et~al\mbox{.}}} \bibinfo{year}{2023}\natexlab{}.
\newblock \showarticletitle{AugGPT: Leveraging ChatGPT for Text Data Augmentation}.
\newblock \bibinfo{journal}{\emph{arXiv preprint arXiv:2302.13007}} (\bibinfo{year}{2023}).
\newblock


\bibitem[Deng et~al\mbox{.}(2022)]%
        {Recent_advances_in_text-to-SQL}
\bibfield{author}{\bibinfo{person}{Naihao Deng}, \bibinfo{person}{Yulong Chen}, {and} \bibinfo{person}{Yue Zhang}.} \bibinfo{year}{2022}\natexlab{}.
\newblock \showarticletitle{Recent Advances in Text-to-SQL: {A} Survey of What We Have and What We Expect}. In \bibinfo{booktitle}{\emph{COLING}}.
\newblock


\bibitem[Dong et~al\mbox{.}(2023)]%
        {dong2023c3}
\bibfield{author}{\bibinfo{person}{Xuemei Dong}, \bibinfo{person}{Chao Zhang}, \bibinfo{person}{Yuhang Ge}, \bibinfo{person}{Yuren Mao}, \bibinfo{person}{Yunjun Gao}, \bibinfo{person}{lu Chen}, \bibinfo{person}{Jinshu Lin}, {and} \bibinfo{person}{Dongfang Lou}.} \bibinfo{year}{2023}\natexlab{}.
\newblock \showarticletitle{C3: Zero-shot Text-to-SQL with ChatGPT}.
\newblock \bibinfo{journal}{\emph{arXiv preprint arXiv:2307.07306}} (\bibinfo{year}{2023}).
\newblock


\bibitem[Gan et~al\mbox{.}(2021)]%
        {natsql}
\bibfield{author}{\bibinfo{person}{Yujian Gan}, \bibinfo{person}{Xinyun Chen}, \bibinfo{person}{Jinxia Xie}, \bibinfo{person}{Matthew Purver}, \bibinfo{person}{John~R. Woodward}, \bibinfo{person}{John Drake}, {and} \bibinfo{person}{Qiaofu Zhang}.} \bibinfo{year}{2021}\natexlab{}.
\newblock \showarticletitle{Natural {SQL}: Making {SQL} Easier to Infer from Natural Language Specifications}. In \bibinfo{booktitle}{\emph{EMNLP}}.
\newblock


\bibitem[Gao et~al\mbox{.}(2023)]%
        {dail-sql}
\bibfield{author}{\bibinfo{person}{Dawei Gao}, \bibinfo{person}{Haibin Wang}, \bibinfo{person}{Yaliang Li}, \bibinfo{person}{Xiuyu Sun}, \bibinfo{person}{Yichen Qian}, \bibinfo{person}{Bolin Ding}, {and} \bibinfo{person}{Jingren Zhou}.} \bibinfo{year}{2023}\natexlab{}.
\newblock \showarticletitle{Text-to-sql empowered by large language models: A benchmark evaluation}.
\newblock \bibinfo{journal}{\emph{arXiv preprint arXiv:2308.15363}} (\bibinfo{year}{2023}).
\newblock


\bibitem[Guo et~al\mbox{.}(2019)]%
        {irnet}
\bibfield{author}{\bibinfo{person}{Jiaqi Guo}, \bibinfo{person}{Zecheng Zhan}, \bibinfo{person}{Yan Gao}, \bibinfo{person}{Yan Xiao}, \bibinfo{person}{Jian{-}Guang Lou}, \bibinfo{person}{Ting Liu}, {and} \bibinfo{person}{Dongmei Zhang}.} \bibinfo{year}{2019}\natexlab{}.
\newblock \showarticletitle{Towards Complex Text-to-SQL in Cross-Domain Database with Intermediate Representation}. In \bibinfo{booktitle}{\emph{ACL}}.
\newblock


\bibitem[Houlsby et~al\mbox{.}(2019)]%
        {adapter}
\bibfield{author}{\bibinfo{person}{Neil Houlsby}, \bibinfo{person}{Andrei Giurgiu}, \bibinfo{person}{Stanislaw Jastrzebski}, \bibinfo{person}{Bruna Morrone}, \bibinfo{person}{Quentin de Laroussilhe}, \bibinfo{person}{Andrea Gesmundo}, \bibinfo{person}{Mona Attariyan}, {and} \bibinfo{person}{Sylvain Gelly}.} \bibinfo{year}{2019}\natexlab{}.
\newblock \showarticletitle{Parameter-Efficient Transfer Learning for {NLP}}. In \bibinfo{booktitle}{\emph{ICML}}.
\newblock


\bibitem[Hu et~al\mbox{.}(2022)]%
        {LoRA}
\bibfield{author}{\bibinfo{person}{Edward~J. Hu}, \bibinfo{person}{Yelong Shen}, \bibinfo{person}{Phillip Wallis}, \bibinfo{person}{Zeyuan Allen{-}Zhu}, \bibinfo{person}{Yuanzhi Li}, \bibinfo{person}{Shean Wang}, \bibinfo{person}{Lu Wang}, {and} \bibinfo{person}{Weizhu Chen}.} \bibinfo{year}{2022}\natexlab{}.
\newblock \showarticletitle{LoRA: Low-Rank Adaptation of Large Language Models}. In \bibinfo{booktitle}{\emph{ICLR}}.
\newblock


\bibitem[Hui et~al\mbox{.}(2022)]%
        {s2sql}
\bibfield{author}{\bibinfo{person}{Binyuan Hui}, \bibinfo{person}{Ruiying Geng}, \bibinfo{person}{Lihan Wang}, \bibinfo{person}{Bowen Qin}, \bibinfo{person}{Yanyang Li}, \bibinfo{person}{Bowen Li}, \bibinfo{person}{Jian Sun}, {and} \bibinfo{person}{Yongbin Li}.} \bibinfo{year}{2022}\natexlab{}.
\newblock \showarticletitle{S{\({^2}\)}SQL: Injecting Syntax to Question-Schema Interaction Graph Encoder for Text-to-SQL Parsers}. In \bibinfo{booktitle}{\emph{ACL}}.
\newblock


\bibitem[Iyer et~al\mbox{.}(2017)]%
        {scholar}
\bibfield{author}{\bibinfo{person}{Srinivasan Iyer}, \bibinfo{person}{Ioannis Konstas}, \bibinfo{person}{Alvin Cheung}, \bibinfo{person}{Jayant Krishnamurthy}, {and} \bibinfo{person}{Luke Zettlemoyer}.} \bibinfo{year}{2017}\natexlab{}.
\newblock \showarticletitle{Learning a Neural Semantic Parser from User Feedback}. In \bibinfo{booktitle}{\emph{ACL}}.
\newblock


\bibitem[Lee et~al\mbox{.}(2021)]%
        {kaggledbqa}
\bibfield{author}{\bibinfo{person}{Chia{-}Hsuan Lee}, \bibinfo{person}{Oleksandr Polozov}, {and} \bibinfo{person}{Matthew Richardson}.} \bibinfo{year}{2021}\natexlab{}.
\newblock \showarticletitle{KaggleDBQA: Realistic Evaluation of Text-to-SQL Parsers}. In \bibinfo{booktitle}{\emph{ACL}}.
\newblock


\bibitem[Lei et~al\mbox{.}(2020)]%
        {lei2020re}
\bibfield{author}{\bibinfo{person}{Wenqiang Lei}, \bibinfo{person}{Weixin Wang}, \bibinfo{person}{Zhixin Ma}, \bibinfo{person}{Tian Gan}, \bibinfo{person}{Wei Lu}, \bibinfo{person}{Min{-}Yen Kan}, {and} \bibinfo{person}{Tat{-}Seng Chua}.} \bibinfo{year}{2020}\natexlab{}.
\newblock \showarticletitle{Re-examining the Role of Schema Linking in Text-to-SQL}. In \bibinfo{booktitle}{\emph{EMNLP}}.
\newblock


\bibitem[Lester et~al\mbox{.}(2021)]%
        {prompt_tuning}
\bibfield{author}{\bibinfo{person}{Brian Lester}, \bibinfo{person}{Rami Al{-}Rfou}, {and} \bibinfo{person}{Noah Constant}.} \bibinfo{year}{2021}\natexlab{}.
\newblock \showarticletitle{The Power of Scale for Parameter-Efficient Prompt Tuning}. In \bibinfo{booktitle}{\emph{EMNLP}}.
\newblock


\bibitem[Li et~al\mbox{.}(2023c)]%
        {RESDSQL}
\bibfield{author}{\bibinfo{person}{Haoyang Li}, \bibinfo{person}{Jing Zhang}, \bibinfo{person}{Cuiping Li}, {and} \bibinfo{person}{Hong Chen}.} \bibinfo{year}{2023}\natexlab{c}.
\newblock \showarticletitle{{RESDSQL:} Decoupling Schema Linking and Skeleton Parsing for Text-to-SQL}. In \bibinfo{booktitle}{\emph{AAAI}}.
\newblock


\bibitem[Li et~al\mbox{.}(2023a)]%
        {graphix-t5}
\bibfield{author}{\bibinfo{person}{Jinyang Li}, \bibinfo{person}{Binyuan Hui}, \bibinfo{person}{Reynold Cheng}, \bibinfo{person}{Bowen Qin}, \bibinfo{person}{Chenhao Ma}, \bibinfo{person}{Nan Huo}, \bibinfo{person}{Fei Huang}, \bibinfo{person}{Wenyu Du}, \bibinfo{person}{Luo Si}, {and} \bibinfo{person}{Yongbin Li}.} \bibinfo{year}{2023}\natexlab{a}.
\newblock \showarticletitle{Graphix-T5: Mixing Pre-trained Transformers with Graph-Aware Layers for Text-to-SQL Parsing}. In \bibinfo{booktitle}{\emph{AAAI}}.
\newblock


\bibitem[Li et~al\mbox{.}(2023b)]%
        {bird}
\bibfield{author}{\bibinfo{person}{Jinyang Li}, \bibinfo{person}{Binyuan Hui}, \bibinfo{person}{Ge Qu}, \bibinfo{person}{Binhua Li}, \bibinfo{person}{Jiaxi Yang}, \bibinfo{person}{Bowen Li}, \bibinfo{person}{Bailin Wang}, \bibinfo{person}{Bowen Qin}, \bibinfo{person}{Rongyu Cao}, \bibinfo{person}{Ruiying Geng}, {et~al\mbox{.}}} \bibinfo{year}{2023}\natexlab{b}.
\newblock \showarticletitle{Can llm already serve as a database interface? a big bench for large-scale database grounded text-to-sqls}. In \bibinfo{booktitle}{\emph{NeurIPS}}.
\newblock


\bibitem[Li and Liang(2021)]%
        {prefix_tuning}
\bibfield{author}{\bibinfo{person}{Xiang~Lisa Li} {and} \bibinfo{person}{Percy Liang}.} \bibinfo{year}{2021}\natexlab{}.
\newblock \showarticletitle{Prefix-Tuning: Optimizing Continuous Prompts for Generation}. In \bibinfo{booktitle}{\emph{ACL}}.
\newblock


\bibitem[Lin et~al\mbox{.}(2022)]%
        {lin2022teaching}
\bibfield{author}{\bibinfo{person}{Stephanie Lin}, \bibinfo{person}{Jacob Hilton}, {and} \bibinfo{person}{Owain Evans}.} \bibinfo{year}{2022}\natexlab{}.
\newblock \showarticletitle{Teaching Models to Express Their Uncertainty in Words}.
\newblock \bibinfo{journal}{\emph{Transactions on Machine Learning Research}} (\bibinfo{year}{2022}).
\newblock


\bibitem[Liu et~al\mbox{.}(2019)]%
        {roberta}
\bibfield{author}{\bibinfo{person}{Yinhan Liu}, \bibinfo{person}{Myle Ott}, \bibinfo{person}{Naman Goyal}, \bibinfo{person}{Jingfei Du}, \bibinfo{person}{Mandar Joshi}, \bibinfo{person}{Danqi Chen}, \bibinfo{person}{Omer Levy}, \bibinfo{person}{Mike Lewis}, \bibinfo{person}{Luke Zettlemoyer}, {and} \bibinfo{person}{Veselin Stoyanov}.} \bibinfo{year}{2019}\natexlab{}.
\newblock \showarticletitle{RoBERTa: {A} Robustly Optimized {BERT} Pretraining Approach}.
\newblock \bibinfo{journal}{\emph{CoRR}}  \bibinfo{volume}{abs/1907.11692} (\bibinfo{year}{2019}).
\newblock


\bibitem[Pourreza and Rafiei(2023)]%
        {DIN-SQL}
\bibfield{author}{\bibinfo{person}{Mohammadreza Pourreza} {and} \bibinfo{person}{Davood Rafiei}.} \bibinfo{year}{2023}\natexlab{}.
\newblock \showarticletitle{DIN-SQL: Decomposed in-context learning of text-to-sql with self-correction}. In \bibinfo{booktitle}{\emph{NeurIPS}}.
\newblock


\bibitem[Qin et~al\mbox{.}(2022)]%
        {Text-to-SQL_survey}
\bibfield{author}{\bibinfo{person}{Bowen Qin}, \bibinfo{person}{Binyuan Hui}, \bibinfo{person}{Lihan Wang}, \bibinfo{person}{Min Yang}, \bibinfo{person}{Jinyang Li}, \bibinfo{person}{Binhua Li}, \bibinfo{person}{Ruiying Geng}, \bibinfo{person}{Rongyu Cao}, \bibinfo{person}{Jian Sun}, \bibinfo{person}{Luo Si}, {et~al\mbox{.}}} \bibinfo{year}{2022}\natexlab{}.
\newblock \showarticletitle{A survey on text-to-sql parsing: Concepts, methods, and future directions}.
\newblock \bibinfo{journal}{\emph{arXiv preprint arXiv:2208.13629}} (\bibinfo{year}{2022}).
\newblock


\bibitem[Raffel et~al\mbox{.}(2020)]%
        {t5}
\bibfield{author}{\bibinfo{person}{Colin Raffel}, \bibinfo{person}{Noam Shazeer}, \bibinfo{person}{Adam Roberts}, \bibinfo{person}{Katherine Lee}, \bibinfo{person}{Sharan Narang}, \bibinfo{person}{Michael Matena}, \bibinfo{person}{Yanqi Zhou}, \bibinfo{person}{Wei Li}, {and} \bibinfo{person}{Peter~J. Liu}.} \bibinfo{year}{2020}\natexlab{}.
\newblock \showarticletitle{Exploring the Limits of Transfer Learning with a Unified Text-to-Text Transformer}.
\newblock \bibinfo{journal}{\emph{Jounal Of Machine Learning Research}}  \bibinfo{volume}{21} (\bibinfo{year}{2020}), \bibinfo{pages}{140:1--140:67}.
\newblock


\bibitem[Rai et~al\mbox{.}(2023)]%
        {Token_Preprocessing}
\bibfield{author}{\bibinfo{person}{Daking Rai}, \bibinfo{person}{Bailin Wang}, \bibinfo{person}{Yilun Zhou}, {and} \bibinfo{person}{Ziyu Yao}.} \bibinfo{year}{2023}\natexlab{}.
\newblock \showarticletitle{Improving Generalization in Language Model-based Text-to-SQL Semantic Parsing: Two Simple Semantic Boundary-based Techniques}. In \bibinfo{booktitle}{\emph{ACL}}.
\newblock


\bibitem[Scholak et~al\mbox{.}(2021)]%
        {picard}
\bibfield{author}{\bibinfo{person}{Torsten Scholak}, \bibinfo{person}{Nathan Schucher}, {and} \bibinfo{person}{Dzmitry Bahdanau}.} \bibinfo{year}{2021}\natexlab{}.
\newblock \showarticletitle{{PICARD:} Parsing Incrementally for Constrained Auto-Regressive Decoding from Language Models}. In \bibinfo{booktitle}{\emph{EMNLP}}.
\newblock


\bibitem[Sun et~al\mbox{.}(2023)]%
        {sqlpalm}
\bibfield{author}{\bibinfo{person}{Ruoxi Sun}, \bibinfo{person}{Sercan~O Arik}, \bibinfo{person}{Hootan Nakhost}, \bibinfo{person}{Hanjun Dai}, \bibinfo{person}{Rajarishi Sinha}, \bibinfo{person}{Pengcheng Yin}, {and} \bibinfo{person}{Tomas Pfister}.} \bibinfo{year}{2023}\natexlab{}.
\newblock \showarticletitle{SQL-PaLM: Improved Large Language ModelAdaptation for Text-to-SQL}.
\newblock \bibinfo{journal}{\emph{arXiv preprint arXiv:2306.00739}} (\bibinfo{year}{2023}).
\newblock


\bibitem[Touvron et~al\mbox{.}(2023)]%
        {llama2}
\bibfield{author}{\bibinfo{person}{Hugo Touvron}, \bibinfo{person}{Louis Martin}, \bibinfo{person}{Kevin Stone}, \bibinfo{person}{Peter Albert}, \bibinfo{person}{Amjad Almahairi}, \bibinfo{person}{Yasmine Babaei}, \bibinfo{person}{Nikolay Bashlykov}, \bibinfo{person}{Soumya Batra}, \bibinfo{person}{Prajjwal Bhargava}, \bibinfo{person}{Shruti Bhosale}, {et~al\mbox{.}}} \bibinfo{year}{2023}\natexlab{}.
\newblock \showarticletitle{Llama 2: Open foundation and fine-tuned chat models}.
\newblock \bibinfo{journal}{\emph{arXiv preprint arXiv:2307.09288}} (\bibinfo{year}{2023}).
\newblock


\bibitem[Wang et~al\mbox{.}(2020)]%
        {ratsql}
\bibfield{author}{\bibinfo{person}{Bailin Wang}, \bibinfo{person}{Richard Shin}, \bibinfo{person}{Xiaodong Liu}, \bibinfo{person}{Oleksandr Polozov}, {and} \bibinfo{person}{Matthew Richardson}.} \bibinfo{year}{2020}\natexlab{}.
\newblock \showarticletitle{{RAT-SQL:} Relation-Aware Schema Encoding and Linking for Text-to-SQL Parsers}. In \bibinfo{booktitle}{\emph{ACL}}.
\newblock


\bibitem[Wei et~al\mbox{.}(2022)]%
        {cot}
\bibfield{author}{\bibinfo{person}{Jason Wei}, \bibinfo{person}{Xuezhi Wang}, \bibinfo{person}{Dale Schuurmans}, \bibinfo{person}{Maarten Bosma}, \bibinfo{person}{Brian Ichter}, \bibinfo{person}{Fei Xia}, \bibinfo{person}{Ed~H. Chi}, \bibinfo{person}{Quoc~V. Le}, {and} \bibinfo{person}{Denny Zhou}.} \bibinfo{year}{2022}\natexlab{}.
\newblock \showarticletitle{Chain-of-Thought Prompting Elicits Reasoning in Large Language Models}. In \bibinfo{booktitle}{\emph{NeurIPS}}.
\newblock


\bibitem[Wu et~al\mbox{.}(2023)]%
        {CoT-KA}
\bibfield{author}{\bibinfo{person}{Dingjun Wu}, \bibinfo{person}{Jing Zhang}, {and} \bibinfo{person}{Xinmei Huang}.} \bibinfo{year}{2023}\natexlab{}.
\newblock \showarticletitle{Chain of Thought Prompting Elicits Knowledge Augmentation}. In \bibinfo{booktitle}{\emph{ACL}}.
\newblock


\bibitem[Xue et~al\mbox{.}(2021)]%
        {mt5}
\bibfield{author}{\bibinfo{person}{Linting Xue}, \bibinfo{person}{Noah Constant}, \bibinfo{person}{Adam Roberts}, \bibinfo{person}{Mihir Kale}, \bibinfo{person}{Rami Al{-}Rfou}, \bibinfo{person}{Aditya Siddhant}, \bibinfo{person}{Aditya Barua}, {and} \bibinfo{person}{Colin Raffel}.} \bibinfo{year}{2021}\natexlab{}.
\newblock \showarticletitle{mT5: {A} Massively Multilingual Pre-trained Text-to-Text Transformer}. In \bibinfo{booktitle}{\emph{NAACL}}.
\newblock


\bibitem[Yang et~al\mbox{.}(2023)]%
        {baichuan2}
\bibfield{author}{\bibinfo{person}{Aiyuan Yang}, \bibinfo{person}{Bin Xiao}, \bibinfo{person}{Bingning Wang}, \bibinfo{person}{Borong Zhang}, \bibinfo{person}{Chao Yin}, \bibinfo{person}{Chenxu Lv}, \bibinfo{person}{Da Pan}, \bibinfo{person}{Dian Wang}, \bibinfo{person}{Dong Yan}, \bibinfo{person}{Fan Yang}, {et~al\mbox{.}}} \bibinfo{year}{2023}\natexlab{}.
\newblock \showarticletitle{Baichuan 2: Open large-scale language models}.
\newblock \bibinfo{journal}{\emph{arXiv preprint arXiv:2309.10305}} (\bibinfo{year}{2023}).
\newblock


\bibitem[Yu et~al\mbox{.}(2018a)]%
        {syntaxsqlnet}
\bibfield{author}{\bibinfo{person}{Tao Yu}, \bibinfo{person}{Michihiro Yasunaga}, \bibinfo{person}{Kai Yang}, \bibinfo{person}{Rui Zhang}, \bibinfo{person}{Dongxu Wang}, \bibinfo{person}{Zifan Li}, {and} \bibinfo{person}{Dragomir Radev}.} \bibinfo{year}{2018}\natexlab{a}.
\newblock \showarticletitle{Syntaxsqlnet: Syntax tree networks for complex and cross-domaintext-to-sql task}.
\newblock \bibinfo{journal}{\emph{arXiv preprint arXiv:1810.05237}} (\bibinfo{year}{2018}).
\newblock


\bibitem[Yu et~al\mbox{.}(2018b)]%
        {spider}
\bibfield{author}{\bibinfo{person}{Tao Yu}, \bibinfo{person}{Rui Zhang}, \bibinfo{person}{Kai Yang}, \bibinfo{person}{Michihiro Yasunaga}, \bibinfo{person}{Dongxu Wang}, \bibinfo{person}{Zifan Li}, \bibinfo{person}{James Ma}, \bibinfo{person}{Irene Li}, \bibinfo{person}{Qingning Yao}, \bibinfo{person}{Shanelle Roman}, \bibinfo{person}{Zilin Zhang}, {and} \bibinfo{person}{Dragomir~R. Radev}.} \bibinfo{year}{2018}\natexlab{b}.
\newblock \showarticletitle{Spider: {A} Large-Scale Human-Labeled Dataset for Complex and Cross-Domain Semantic Parsing and Text-to-SQL Task}. In \bibinfo{booktitle}{\emph{EMNLP}}.
\newblock


\bibitem[Zelle and Mooney(1996)]%
        {genquery}
\bibfield{author}{\bibinfo{person}{John~M. Zelle} {and} \bibinfo{person}{Raymond~J. Mooney}.} \bibinfo{year}{1996}\natexlab{}.
\newblock \showarticletitle{Learning to Parse Database Queries Using Inductive Logic Programming}. In \bibinfo{booktitle}{\emph{AAAI}}.
\newblock


\bibitem[Zhao et~al\mbox{.}(2023)]%
        {zhao2023survey}
\bibfield{author}{\bibinfo{person}{Wayne~Xin Zhao}, \bibinfo{person}{Kun Zhou}, \bibinfo{person}{Junyi Li}, \bibinfo{person}{Tianyi Tang}, \bibinfo{person}{Xiaolei Wang}, \bibinfo{person}{Yupeng Hou}, \bibinfo{person}{Yingqian Min}, \bibinfo{person}{Beichen Zhang}, \bibinfo{person}{Junjie Zhang}, \bibinfo{person}{Zican Dong}, {et~al\mbox{.}}} \bibinfo{year}{2023}\natexlab{}.
\newblock \showarticletitle{A survey of large language models}.
\newblock \bibinfo{journal}{\emph{arXiv preprint arXiv:2303.18223}} (\bibinfo{year}{2023}).
\newblock


\bibitem[Zhong et~al\mbox{.}(2020)]%
        {test_suite}
\bibfield{author}{\bibinfo{person}{Ruiqi Zhong}, \bibinfo{person}{Tao Yu}, {and} \bibinfo{person}{Dan Klein}.} \bibinfo{year}{2020}\natexlab{}.
\newblock \showarticletitle{Semantic Evaluation for Text-to-SQL with Distilled Test Suites}. In \bibinfo{booktitle}{\emph{EMNLP}}.
\newblock


\bibitem[Zhong et~al\mbox{.}(2017)]%
        {wikisql}
\bibfield{author}{\bibinfo{person}{Victor Zhong}, \bibinfo{person}{Caiming Xiong}, {and} \bibinfo{person}{Richard Socher}.} \bibinfo{year}{2017}\natexlab{}.
\newblock \showarticletitle{Seq2sql: Generating structured queries from natural language using reinforcement learning}.
\newblock \bibinfo{journal}{\emph{arXiv preprint arXiv:1709.00103}} (\bibinfo{year}{2017}).
\newblock


\end{thebibliography}









\end{document}